\documentclass[journal]{IEEEtran}
\usepackage{amsmath,amsfonts}
\usepackage{algorithmic}
\usepackage{algorithm}
\usepackage{array}
\usepackage{textcomp}
\usepackage{stfloats}
\usepackage[hyphens]{url}
\usepackage{verbatim}
\usepackage{graphicx}
\usepackage{cite}
\usepackage{xcolor}
\usepackage{bbm}
\usepackage{colortbl}
\definecolor{mygray}{gray}{.9}
\definecolor{mypink}{rgb}{.99,.91,.95}
\definecolor{mycyan}{cmyk}{.3,0,0,0}

\usepackage{adjustbox}
\usepackage{array,booktabs}
\usepackage{physics}

\newcolumntype{C}[1]{>{\raggedright\arraybackslash}p{#1}}

\usepackage{tikz}
\usepackage[edges]{forest}
\definecolor{hidden-draw}{RGB}{205, 44, 36}
\definecolor{hidden-blue}{RGB}{194,232,247}
\definecolor{hidden-orange}{RGB}{243,202,120}
\definecolor{hidden-yellow}{RGB}{242,244,193}
\definecolor{tree-level-1}{RGB}{245,20,85}
\definecolor{tree-level-2}{RGB}{246,86,118}
\definecolor{tree-level-3}{RGB}{248,177,193}
\definecolor{tree-leaf}{RGB}{176,230,198}

\definecolor{Self}{RGB}{255,0,128}
\definecolor{Ensemble}{RGB}{0,127,255}
\definecolor{Iterative}{RGB}{153,51,255}

\definecolor{exemplar1}{RGB}{136,98,148}
\definecolor{exemplar2}{RGB}{148,210,242}
\definecolor{knowledge1}{RGB}{249,219,152}
\definecolor{knowledge2}{RGB}{255,245,220}

\usepackage{pgfplots}
\pgfplotsset{compat=1.17}
\usetikzlibrary{pgfplots.groupplots}

\usepackage{threeparttable}

\usepackage[hidelinks]{hyperref}
\hypersetup{
   linkcolor=blue,
   breaklinks=true,   
   colorlinks=true,   
}

\begin{document}
\title{Foundation Models in Robotics: Applications, Challenges, and the Future}
\author{Roya Firoozi$^1$, Johnathan Tucker$^1$, Stephen Tian$^1$, Anirudha Majumdar$^{2,6}$, Jiankai Sun$^1$, \\ Weiyu Liu$^1$, Yuke Zhu$^{3,4}$,
Shuran Song$^1$,
Ashish Kapoor$^5$, Karol Hausman$^{1,6}$, \\
Brian Ichter$^6$, Danny Driess$^{6,7}$, 
Jiajun Wu$^1$, Cewu Lu$^8$, Mac Schwager$^1$\\ \quad
\\
$^1$Stanford University, $^2$Princeton University, $^3$UT Austin, $^4$NVIDIA, 
$^5$Scaled Foundations, \\
$^6$Google DeepMind, $^7$TU Berlin, $^8$Shanghai Jiao Tong University
}

\markboth{}%
{}


\maketitle

\tableofcontents
\clearpage
\newpage

\begin{abstract}
We survey applications of pretrained foundation models in robotics. Traditional deep learning models in robotics are trained on small datasets tailored for specific tasks, which limits their adaptability across diverse applications. In contrast, foundation models pretrained on internet-scale data appear to have superior generalization capabilities, and in some instances display an emergent ability to find zero-shot solutions to problems that are not present in the training data. Foundation models may hold the potential to enhance various components of the robot autonomy stack, from perception to decision-making and control. For example, large language models can generate code or provide common sense reasoning, while vision-language models enable open-vocabulary visual recognition. However, significant open research challenges remain, particularly around the scarcity of robot-relevant training data, safety guarantees and uncertainty quantification, and real-time execution.  In this survey, we study recent papers that have used or built foundation models to solve robotics problems. We explore how foundation models contribute to improving robot capabilities in the domains of perception, decision-making, and control. We discuss the challenges hindering the adoption of foundation models in robot autonomy and provide opportunities and potential pathways for future advancements. 
The GitHub project corresponding to this paper\footnote{Preliminary release. We are committed to further enhancing and updating this work to ensure its quality and relevance} can be found \href{https://github.com/robotics-survey/Awesome-Robotics-Foundation-Models}{here}.

\end{abstract}


\section{Introduction}\label{SECI}
\IEEEPARstart{F}{oundation} models are pretrained on extensive internet-scale data and can be fine-tuned for adaptation to a wide range of downstream tasks. Foundation models have demonstrated significant breakthroughs in vision and language processing; examples include BERT~\cite{devlin2018bert}, GPT-3~\cite{brown2020language}, GPT-4~\cite{openai2023gpt4}, CLIP~\cite{CLIP2021}, DALL-E~\cite{ramesh2021zero}, and PaLM-E~\cite{driess2023palme}. Foundation models have the potential to unlock new possibilities in robotics domains such as autonomous driving, household robotics, industrial robotics, assistive robotics, medical robotics, field robotics, and multi-robot systems. Pretrained Large Language Models (LLMs), Large Vision-Language Models (VLMs), Large Audio-Language Models (ALMs), and Large Visual-Navigation Models (VNMs) can be utilized to improve various tasks in robotics settings. The integration of foundation models into robotics is a rapidly evolving area, and the robotics community has very recently started exploring ways to leverage these large models within the robotics domain for perception, prediction, planning, and control. 

Prior to the emergence of foundation models, traditional deep learning models for robotics were typically trained on limited datasets gathered for distinct tasks~\cite{sun2022plate}. Conversely, foundation models are pre-trained on extensive and diverse data, which has been proven in other domains (such as natural language processing, computer vision, and healthcare~\cite{qiu2023large}) to significantly expand adaptability, generalization capability, and overall performance. Ultimately, foundation models may hold the potential to yield these same benefits in robotics. 
Knowledge transfer from foundation models may reduce training time and computational resources compared to task-specific models. Particularly relevant to robotics, multimodal foundation models can fuse and align multimodal heterogeneous data gathered from various sensors into compact homogeneous representations needed for robot understanding and reasoning~\cite{reasoning2023book}. These learned representations hold the potential to be used in any part of the autonomy stack including perception, decision-making, and control. Furthermore, foundation models provide zero-shot capabilities, which refer to the ability of an AI system to perform tasks without prior examples or dedicated training data for that specific task. The would enable robots to generalize their learned knowledge to novel cases, enhancing adaptability and flexibility for robots in unstructured settings.  

Integrating foundation models into robotic systems may enable context-aware robotic systems by enhancing the robot's ability to perceive and interact with the environment. For example in the perception domain, Large Vision-Language Models (VLMs) have been found to provide cross-modal understanding by learning associations between visual and textual data, aiding tasks such as zero-shot image classification, zero-shot object detection~\cite{zhang2023sam3d}, and 3D classification~\cite{3dllm}. As another example, language grounding in the 3D world \cite{chen2022leveraging} (aligning contextual understanding of VLMs to the 3-dimensional (3D) real world) may enhance a robot's spatial awareness by associating words with specific objects, locations, or actions within the 3D environment.

In the decision-making or planning domain, LLMs and VLMs have been found to assist robots in task specification for high-level planning \cite{FoundationModelSurvey2023}. 
Robots can perform more complex tasks by leveraging linguistic cues in manipulation, navigation, and interaction. For example, for robot policy learning techniques like imitation learning \cite{mandi2022cacti} and reinforcement learning \cite{palo2023towards}, foundation models seem to offer the possibility to improve data efficiency and enhance contextual understanding. In particular, language-driven rewards can be used to guide RL agents by providing shaped rewards\cite{kwon2023reward}. Also, researchers have employed language models to provide feedback for policy learning techniques \cite{feng2023chessgpt}. 
Some works have shown that a VLM model's visual question-answering (VQA) capability can be harnessed in robotics use cases. For example, researchers have used VLMs to answer questions related to visual content to aid robots in accomplishing their tasks \cite{du2023zeroshot}. Also, researchers have stated utilizing VLMs to help with data annotation, by generating descriptive labels for visual content \cite{he2023annollm}.  

Despite the transformative capabilities of foundation models in vision and language processing, the generalization and fine-tuning of foundation models for real-world robotics tasks remain challenging. These challenges include: 1) \textbf{Data Scarcity:} how to obtain internet-scale data for robot manipulation, locomotion, navigation, and other robotics tasks, and how to perform self-supervised training with this data, 2) \textbf{High Variability:} how to deal with the large diversity in physical environments, physical robot platforms, and potential robot tasks while still maintaining the generality required for a foundation model,  3) \textbf{Uncertainty Quantification:} how to deal with (i) instance-level uncertainty such as language ambiguity or LLM hallucination; (ii) distribution-level uncertainty; and (iii) distribution-shift, especially resulting from closed-loop robot deployment, 4) \textbf{Safety Evaluation:} How to rigorously test for the safety of a foundation model-based robotic system (i) prior to deployment, (ii) as the model is updated throughout its lifecycle, and (iii) as the robot operates in its target environments. 5) \textbf{Real-Time Performance: }how to deal with the high inference time of some foundation models which could hinder their deployment on robots and how to accelerate inference in foundation models to the speed required for online decision-making.  

In this survey, we study the existing literature on the use of foundation models in robotics. We study current approaches and applications, present current challenges, suggest directions for future research to address these challenges, and identify potential risks exposed by integrating foundation models into robot autonomy. Another survey on foundation models in robotics appeared simultaneously with ours on arXiv \cite{xiao2023robot}. In comparison with that paper, ours emphasizes future challenges and opportunities, including safety and risk, and ours has a stronger emphasis on comparisons in applications, algorithms, and architectures among the existing papers in this space.  In contrast to some existing surveys that focus on a specific in-context instruction, such as prompts~\cite{Liu2023a}, vision transformers~\cite{VisionTransformerSurvey}, or decision-making~\cite{FoundationModelSurvey2023}, \cite{Muning}, we provide a broader perspective to connect distinct research threads in foundation models organized around their relevance to and application to robotics. Conversely, our scope is much narrower than the paper \cite{bommasani2021opportunities}, which explores the broad application of foundation models across many disciplines, of which robotics is one. We hope this paper can provide clarity regarding areas of recent progress and existing deficiencies in the research, and point the way forward to future opportunities and challenges facing this research area.  Ultimately, we aim to give a resource for robotics researchers to learn about this exciting new area.

We limit the scope of this survey to papers that fall into one of the following categories: 
\begin{enumerate}
    \item{\bf Background Papers:} Papers that do not explicitly link to robotics, but are nonetheless required for understanding foundation models. These papers are discussed in the background section (section \ref{Foundation Models Background}) of the survey paper.

    \item{\bf Robotics Papers:}  Papers that integrate a foundation model into a robotic system in a plug-and-play fashion, papers that adapt or fine-tune foundation models for robotic systems, or papers that build new robotic-specific foundation models. 
    
    \item{\bf Robotics-Adjacent Papers:} Papers that present methods or techniques applied to areas adjacent to robotics (e.g., computer vision, embodied AI), with a clear path to future application in robotics.
\end{enumerate}

This survey is organized as follows: In Section~\ref{Foundation Models Background}, we provide an introduction to foundation models including LLMs, vision transformers, VLMs, embodied multimodal language models, and visual generative models. In addition, in the last part of this section, we discuss different training methods used to train foundation models. In Section \ref{Decision-Making}, we present a review of how foundation models are integrated into different tasks for decision-making in robotics. First, we discuss robot policy learning using language-conditioned imitation learning, and language-assisted reinforcement learning. Then, we discuss how to use foundation models to design a language-conditioned value function that can be used for planning purposes. Next, robot task specification and code generation for task planning using foundation models are presented. In Section~\ref{Perception}, we study various perception tasks in robotics that have the potential to be enhanced by employing foundation models. These tasks include semantic segmentation, 3D scene representation, zero-shot 3D classification, affordance prediction, and dynamics prediction. In Section \ref{EmbodiedAI}, we present papers about Embodied AI agents, generalist AI agents, as well as simulators and benchmarks developed for embodied AI research. In Section \ref{challenges and Future Directions}, we conclude the survey by discussing different challenges for employing foundation models in robotic systems and proposing potential avenues for future research. Finally, in Section \ref{Conclusion} we offer the concluding remarks.  

\tikzstyle{my-box}=[
    rectangle,
    draw=hidden-draw,
    rounded corners,
    text opacity=1,
    minimum height=1.5em,
    minimum width=5em,
    inner sep=2pt,
    align=center,
    fill opacity=.5,
]
\tikzstyle{leaf}=[my-box, minimum height=1.5em,
    fill=white!70!red, text=black, align=left,font=\scriptsize,
    inner xsep=2pt,
    inner ysep=4pt,
]
\begin{figure*}[t]
    \centering
    \resizebox{\textwidth}{!}
    {
        \begin{forest}
            forked edges,
            for tree={
                grow=east,
                reversed=true,
                anchor=base west,
                parent anchor=east,
                child anchor=west,
                base=left,
                font=\small,
                rectangle,
                draw=hidden-draw,
                rounded corners,
                align=left,
                minimum width=4em,
                edge={darkgray, line width=1pt},
                inner xsep=2pt,
                inner ysep=3pt},
                ver/.style={rotate=90, child anchor=north, parent anchor=south, anchor=center},
                where level=1{text width=6em,font=\scriptsize,}{},
                where level=2{text width=7em,font=\scriptsize,}{},
                where level=3{text width=7em,font=\scriptsize,}{},
                where level=4{text width=8em,font=\scriptsize,}{},
            [Foundation Models in Robotics, ver, 
                [Robotics, ver,
                    [Robot Policy Learning
                    [Language-Conditioned \\ Imitation Learning
                        [{e.g. CLIPort~\cite{shridhar2021cliport}, Play-LMP~\cite{lynch2019play}, PerAct~\cite{shridhar2022peract},
                            Multi-Context Imitation~\cite{lynch2021language}, \\CACTI~\cite{mandi2022cacti}, Voltron~\cite{karamcheti2023language}}, leaf, text width=25em]
                    ]
                    [Language-Assisted \\ Reinforcement Learning
                        [{e.g. Adaptive Agent (AdA)~\cite{adaptiveagentteam2023humantimescale}, Palo et al.~\cite{palo2023towards}}, leaf, text width=25em]
                    ]
                    ]
                    [Language-Image \\ Goal-Conditioned \\ Value Learning
                        [{e.g. R3M~\cite{nair2022r3m}, SayCan~\cite{Ahn2022DoAI}, Inner Monologue~\cite{huang2022inner}, VoxPoser~\cite{VoxPoser2023}, Mahmoudieh et al.~\cite{mahmoudieh2022zeroshot}, VIP~\cite{VIP2022},\\ LIV~\cite{ma2023liv}, LOREL~\cite{nair2022a}}, leaf, text width=30em]
                   ]
                   [High-Level \\ Task Planning
                        [{e.g. NL2TL~\cite{chen2023nl2tl}, Chen et al.~\cite{Chen2023}}, leaf, text width=30em]
                   ]
                    [LLM-Based \\ Code Generation
                        [{e.g. ProgPrompt~\cite{progprompt2023}, Code-as-Policy~\cite{codeaspolicies2022}, ChatGPT-Robotics \cite{vemprala2023chatgpt}}, leaf, text width=30em]
                    ]
                    [Robot Transformers
                    [{e.g. RT-1~\cite{RT12023}, RT-2~\cite{rt22023arxiv}, RT-X~\cite{padalkar2023rtx}, PACT~\cite{Bonatti2023}, Xiao et al.~\cite{Xiao2022}, Radosavovic et al.~\cite{Radosavovic2022}, LATTE~\cite{Bucker2023}}, leaf, text width=30em]
                    ]
                ]
                [Relevant to Robotics, ver
                [Perception 
                    [Open-Vocabulary \\ Object Detection \\ and 3D classification 
                        [{e.g. OWL-ViT~\cite{minderer2022simple},
                        GLIP~\cite{GLIP2022}, Grounding DINO~\cite{liu2023grounding}, \\
                        PointCLIP~\cite{PointCLIP2022},
                        PointBERT \cite{yu2021pointbert},
                        ULIP~\cite{xue2022ulip,xue2023ulip2} },leaf, text width=25em]
                    ]
                    [Open-Vocabulary \\ Semantic Segmentation 
                        [{e.g. LSeg~\cite{li2022languagedriven}, Segment Anything~\cite{kirillov2023segment}, FastSAM~\cite{zhao2023fast}, MobileSAM~\cite{mobile_sam},\\ Track Anything Model (TAM)~\cite{yang2023track}}, leaf, text width=25em]
                    ]
                    [Open-Vocabulary 3D \\Scene Representation
                        [{e.g. CLIP-NERF~\cite{wang2022clip}, LERF~\cite{lerf2023}, DFF~\cite{kobayashi2022decomposing}},leaf, text width=25em]
                    ]
                    [Affordances
                        [{e.g. Affordance Diffusion~\cite{ye2023affordance}, VRB~\cite{bahl2023affordances}},leaf, text width=25em]
                    ]
                ]
                [Embodied AI
                        [{e.g. Huang et al.~\cite{huang2022language}, Statler~\cite{yoneda2023statler}, EmbodiedGPT~\cite{mu2023embodiedgpt}, MineDojo~\cite{fan2022minedojo}, VPT~\cite{baker2022video}, Kwon et al.~\cite{kwon2023reward},\\
                        Voyager~\cite{wang2023voyager}, ELLM~\cite{GudingRL2023}},leaf, text width=33.5em]
                ]
                ]
            ]
        \end{forest}
    }
    \caption{Overview of Robotics Tasks Leveraging Foundation Models.}
    \label{categorization_of_reasoning}
\end{figure*}
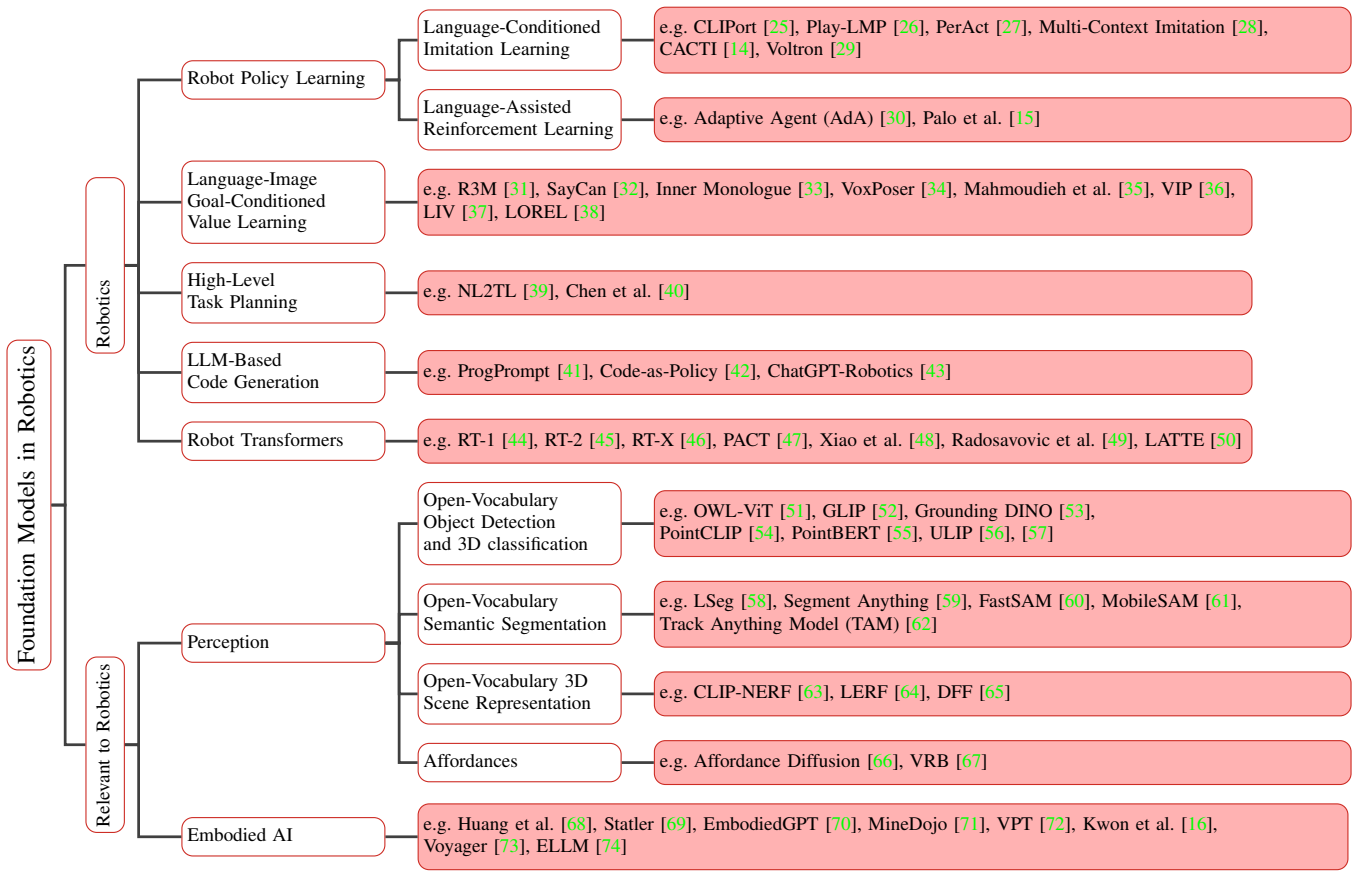


    

    







\section{Foundation Models Background}\label{Foundation Models Background}
Foundation models have billions of parameters and are pretrained on massive internet-scale datasets. Training models of such scale and complexity involve substantial costs. Acquiring, processing, and managing data can be costly. The training process demands significant computational resources, requiring specialized hardware such as GPUs or TPUs, as well as software and infrastructure for model training which requires financial resources. Additionally, training a foundation model is time-intensive, which can translate to even higher costs. Hence these models are often used as plug-and-play modules (which refers to the integration of foundation models into various applications without the need for extensive customization). Table \ref{table:1} provides details about commonly used foundation models. In the rest of this section, we introduce LLMs, vision transformers, VLMs, embodied multi-modal language models, and visual generative models. In the last part of this section, we introduce different training methods that are used to train foundation models.      
\subsection{Terminology and Mathematical Preliminaries}
In this section, we first introduce common terminologies in the context of foundation models and describe basic mathematical details and training practices for various types of foundation models. 

\textbf{Tokenization: }
Given a sequence of characters, tokenization is the process of dividing the sequence into smaller units, called tokens. 
Depending on the tokenization strategy, tokens can be characters, segments of words, complete words, or portions of sentences. Tokens are represented as 1-hot vectors of dimension equal to the size of the total vocabulary and are mapped to lower-dimensional vectors of real numbers through a learned embedding matrix. An LLM takes a sequence of these embedding vectors as raw input, producing a sequence of embedding vectors as raw output.  These output vectors are then mapped back to tokens and hence to text. GPT-3, for example, has a vocabulary of 50,257 different tokens, and an embedding dimension of 12,288.  

The token decoding (from low-dimension real-valued embedding vectors to high-dimension 1-hot vectors) is not deterministic, resulting in a weighting for each possible token in the vocabulary.  These weightings are often used by LLMs as probabilities over tokens, to introduce randomness in the text generation process. For example, the temperature parameter in GPT-3 blends between always choosing the top-weighted token (temperature of 0) and drawing the token based on the probability distribution suggested by the weights (temperature of 1).  This source of randomness is only in the token decoding process, not in the LLM itself.  To the authors' knowledge, this is, in fact, the only source of randomness in the GPT family of models.

One of the most common tokenization schemes, which is used by the GPT family of models, is called byte-pair encoding \cite{sennrich2015neural}.  Byte-pair encoding starts with a token for each individual symbol (e.g., letter, punctuation), then recursively builds tokens by grouping pairs of symbols that commonly appear together, building up to assign tokens to larger and larger groups (pairs of pairs, etc) that appear frequently together in a text corpus.  The tokenization process can extend beyond text data to diverse contexts, encompassing various data modalities like images, videos, and robot actions. In these scenarios, the respective data modalities can be treated as sequential data and tokenized similarly to train generative models. For example, just as language constitutes a sequence of words, an image comprises a sequence of image patches, force sensors yield a sequence of sensory inputs at each time step, and a series of actions represent the sequential nature of tasks for a robot. 

\textbf{Generative Models: }A generative model is a model that learns to sample from a probability distribution to create examples of data that seem to be from the same distribution as the training data.  For example, a face generation model can produce images of faces that cannot be distinguished from the set of real images used to train the model.  These models can be trained to be conditional, meaning they generate samples from a conditional distribution conditioned on a wide range of possible conditioning information. For example, a gender conditional face generator can generate images of female or male faces, where the desired gender is given as a conditioning input to the model. 

\textbf{Discriminative Models: }Discriminative models are used for regression or classification tasks. In contrast to generative models, discriminative models are trained to distinguish between different classes or categories. Their emphasis lies in learning the boundaries between classes within the input space. While generative models learn to sample from the distribution over the data, discriminative models learn to evaluate the probability distribution of the output labels given the input features, or (depending on how the model is trained) learn to evaluate some statistic of the probability distribution over the outputs, such as the expected output given an input.
 
\textbf{Transformer Architecture: }Most foundation models are built on the transformer architecture, which has been instrumental in the rise of foundation models and large language models.  The following discussion was synthesized from \cite{vaswani2017attention}, as well as online blogs, unpublished reports, and wikipedia \cite{DugasGPTBlog,WikipediaGPT3,ThickstunTransformerReport}. A transformer acts simultaneously on a collection of embedded token vectors $(x_1, \ldots, x_N)$ known as a context window.  The key enabling innovation of the Transformer architecture is the multi-head self-attention mechanism originally proposed in the seminal work~\cite{vaswani2017attention}.  In this architecture, each attention head computes a vector of importance weights that corresponds to how strongly a token in the context window $x_i$ correlates with other tokens in the same window $x_j$. Each attention head mathematically encodes different notions of similarity, through different projection matrices used in the computation of the importance weights.  Each head can be trained (backward pass) and evaluated (forward pass) in parallel across all tokens and across all heads, leading to faster training and inference when compared with previous models based on RNNs or LSTMs.  

Mathematically, an attention head maps each token $x_i$ in the context window to a ``query" $q_i = W_q^Tx_i$, and each other token in the context head $x_j$ to a ``key" $k_j = W^T_kx_j$.  The similarity between query and key is then measured through a scaled dot product, $q_i^Tk_j/\sqrt{d}$, where $d$ is the dimension of the query and key vectors. 
 A softmax is then taken over all $j$ to give weights $\alpha_{ij}$ representing how much $x_i$ ``attends to'' $x_j$.  The tokens are then mapped to ``values" with $v_j = W_v^Tx_j$, and the output of the attention for position $i$ is then given as a sum over values weighted by attention weights, $\sum_j\alpha_{ij}v_j$.  One of the key reasons for the success of the transformer attention model is that it can be efficiently computed with GPUs and TPUs by parallelizing the preceding steps into matrix computations,
\begin{align}
    \operatorname{attn}(\mathbf{Q}, \mathbf{K}, \mathbf{V})=\operatorname{softmax}\left(\frac{\mathbf{Q} \mathbf{K}^{\top}}{\sqrt{d_k}}\right) \mathbf{V},
    \label{eq:attn}
\end{align}
where $\mathbf{Q}$, $\mathbf{K}$, $\mathbf{V}$ are matrices with rows $q_i^T$, $k_i^T$, and $v_i^T$, respectively. Each head in the model produces this computation independently, with different $W_q, W_k, W_v$ matrices to encode different kinds of attention.  The outputs from each head are then concatenated, normalized with a skip connection, passed through a fully connected ReLU layer, and normalized again with a skip connection to produce the output of the attention layer. Multiple layers are arranged in various ways to give ``encoders'' and ``decoders,'' which together make up a transformer. 

The size of a transformer model is typically quantified by (i) the size of the context window, (ii) the number of attention heads per layer, (iii) the size of the attention vectors in each head, and (iii) the number of stacked attention layers.  For example, GPT-3's context window is 2048 tokens (corresponding to about 1500 words of text), each attention layer has 96 heads, each head has attention vectors of 128 dimensions, and there are 96 stacked attention layers in the model.

The basic multi-head attention mechanism does not impose any inherent sense of sequence or directionality in the data. However, transformers---especially in natural language applications---are often used as sequence predictors by imposing a positional encoding on the input token sequence. They are then applied to a token sequence autoregressively, meaning they predict the next token in the sequence, add that token to their context window, and repeat. This concept is elaborated below.

\textbf{Autoregressive Models:} The concept of autoregression has been applied in many fields as a representation of random processes whose outputs depend causally on the previous outputs. Autoregressive models use a window of past data to predict the next data point in a sequence. The window then slides one position forward, recursively ingesting the predicted data point into the window and expelling the oldest data point from the window.  The model again predicts the next data point in the sequence, repeating this process indefinitely.  Classical linear autoregressive models such as Auto-Regressive Moving Average (ARMA) and Auto-Regressive Moving Average with eXogenous input (ARMAX) models are standard statistical tools dating back to at least the 1970s \cite{box2015time}.  These modeling concepts were adapted to deep learning models first with RNNs, and later LSTMs, which are both types of learnable nonlinear autoregressive models.  Transformer models, although they are not inherently autoregressive, are often adapted to an autoregressive framework for text prediction tasks.

For example, the GPT family \cite{radford2018improving} builds on the original transformer model by using a modification introduced in \cite{liu2018generating} that removes the transformer encoder blocks entirely, retaining just the transformer decoder blocks. This has the advantage of reducing the number of model parameters by close to half while reducing redundant information that is learned in both the encoder and decoder. During training, the GPT model seeks to produce an output token from the tokenized corpus $\mathcal{X} = (x_1,...,x_n)$ to minimize the negative log-likelihood within the context window of length $N$,

\begin{align}
    \mathcal{L}_{\mathrm{LLM}}=-\sum_i \log P\left(x_i \mid x_{i-N}, \ldots, x_{i-1}\right).
    \label{eq:gpt}
\end{align}
This results in a large pretrained model that autoregressively predicts the next likely token given the tokens in the context window. Although powerful, the unidirectional autoregressive nature of the GPT family means that these models may lag in performance on bidirectional tasks such as reading comprehension. 

\textbf{Masked Auto-Encoding: }To address the unidirectional limitation of the GPT family and allow the model to make bidirectional predictions, works such as BERT \cite{devlin2018bert} use masked auto-encoding. This is achieved through an architectural change, namely the addition of a bidirectional encoder, as well as a novel pre-training objective known as masked language modeling (MLM). The MLM task simply masks a percentage of the tokens in the corpus and requires the model to predict these tokens. Through this procedure, the model is encouraged to learn the context that surrounds a word rather than just the next likely word in a sequence.

\textbf{Contrastive Learning:} Visual-language foundation models such as CLIP \cite{CLIP2021} typically rely on different training methods from the ones used with large language models which encourage explicitly predictive behavior. Visual-language models use contrastive representation learning, where the goal is to learn a joint embedding space between input modalities where similar sample pairs are closer than dissimilar ones. The training objective for many VLMs is some variation of the objective function, 
\begin{align}
    \ell_i^{(v \rightarrow u)}=-\log \frac{\exp \left(\operatorname{sim}\left(\mathbf{v}_i, \mathbf{u}_i)\right / \tau\right)}{\sum_{k=1}^N \exp \left(\operatorname{sim}\left(\mathbf{v}_i, \mathbf{u}_k)\right / \tau\right)},\\
    \ell_i^{(u \rightarrow v)}=-\log \frac{\exp \left(\operatorname{sim}\left(\mathbf{u}_i, \mathbf{v}_i)\right / \tau\right)}{\sum_{k=1}^N \exp \left(\operatorname{sim}\left(\mathbf{u}_i, \mathbf{v}_k)\right / \tau\right)},\\
    \mathcal{L}=\frac{1}{N} \sum_{i=1}^N\left(\lambda \ell_i^{(v \rightarrow u)}+(1-\lambda) \ell_i^{(u \rightarrow v)}\right).
    \label{eq:simclr}
\end{align}
This objective function was popularized for multimodal input by ConVIRT \cite{zhang2022contrastive} and first presented in prior works \cite{chen2020simple,sohn2016improved, wu2018unsupervised, oord2018representation}. This objective function trains the image and text encoders to preserve mutual information between the true text and image pairs. In these equations, $\mathbf{u}_i$ and $\mathbf{v}_i$ are the $i^{\text{th}}$ encoded text and image respectively from $i \in 1,...,N$ image and text pairs. The $\operatorname{sim}$ operation is the cosine similarity between the text and image embeddings, and $\tau$ is a temperature term. In CLIP \cite{CLIP2021} the authors use a symmetric cross-entropy loss, meaning the final loss is an average of the two loss components where each is equally weighted (i.e. $\lambda = 0.5$). 

\textbf{Diffusion Models:} Outside of large language models and multi-modal models such as VLMs, diffusion models for image generation (e.g. DALL-E2) \cite{DALE2} are another class of foundation models considered in this survey. Although diffusion models were established in prior work \cite{pmlr-v37-sohl-dickstein15,song2019generative} the diffusion probabilistic model presented in \cite{ho2020denoising} popularized the method. The diffusion probabilistic model is a deep generative model that is trained in an iterative forward and reverse process. The forward process adds Gaussian noise to an input $x_0$ in a Markov chain until $x_T$ when the result is zero mean isotropic noise. This means the forward process produces a trajectory of noise $q(x_{1:T}|x_0)$ as,
\begin{align}
    q\left(\mathbf{x}_{1: T} \mid \mathbf{x}_0\right):=\prod_{t=1}^T q\left(\mathbf{x}_t \mid \mathbf{x}_{t-1}\right).
    \label{eq:fordiff}
\end{align}
At each time step $q(x_t|x_{t-1})$ is described by a normal distribution with mean $\sqrt{1-\beta_t}\mathbf{x}_{t-1}$ and covariance $\beta_t\mathbf{I}$ where $\beta_t$ is scheduled or a fixed hyperparameter.

The reverse process requires the model to learn to the transitions that will de-noise the zero mean Gaussian and produce the input image. This process is also defined as a Markov chain where the transition distribution at time $t$ is $p_\theta\left(\mathbf{x}_{t-1} \mid \mathbf{x}_t\right):=\mathcal{N}\left(\mathbf{x}_{t-1} ; \boldsymbol{\mu}_\theta\left(\mathbf{x}_t, t\right), \mathbf{\Sigma}_\theta\left(\mathbf{x}_t, t\right)\right)$. For completeness, the reverse process Markov chain is given by, 
\begin{align}
    p_\theta\left(\mathbf{x}_{0: T}\right):=p\left(\mathbf{x}_T\right) \prod_{t=1}^T p_\theta\left(\mathbf{x}_{t-1} \mid \mathbf{x}_t\right).
    \label{eq:revdiff}
\end{align}
Diffusion models are trained using a reduced form of the evidence lower bound loss function that is typical of variational generative models like variational autoencoders (VAEs). The reduced loss function used for training is
\begin{align}
    \label{eq:diffloss}
    \mathcal{L} &= \mathbb{E}_q[D_{\mathrm{KL}}(q(\mathbf{x}_T \mid \mathbf{x}_0) \| p(\mathbf{x}_T))\\
    &+\sum_{t>1} D_{\mathrm{KL}}(q(\mathbf{x}_{t-1} \mid \mathbf{x}_t, \mathbf{x}_0) \| p_\theta(\mathbf{x}_{t-1} \mid \mathbf{x}_t) \nonumber\\
    &-\log p_\theta(\mathbf{x}_0 \mid \mathbf{x}_1)]\nonumber,
\end{align}
where $D_{\mathrm{KL}}(q||p)$ denotes Kullback–Leibler divergence, which is a measure of how different a distribution $q$ is from a distribution $p$.

\subsection{Large Language Model (LLM) Examples and Historical Context}
LLMs have billions of parameters and are trained on trillions of tokens. 
This large scale has allowed models such as GPT-2 \cite{radford2019language} and BERT \cite{devlin2018bert} to achieve state-of-the-art performance on the Winograd Schema challenge \cite{levesque2012winograd} and the General Language Understanding Evaluation (GLUE) \cite{wang2018glue} benchmarks, respectively. Their successors include GPT-3 \cite{brown2020language}, LLaMA \cite{touvron2023llama}, and PaLM \cite{Chowdhery2022PaLMSL} has grown considerably in the number of parameters (typically now over 100 billion), the size of the context window (typically now over 1000 tokens), and the size of the training data set (typically now 10s of terabytes of text). GPT-3 is trained on the Common Crawl dataset. Common Crawl contains petabytes of publicly available data over 12 years of web crawling and includes raw web page data, metadata, and text extracts. LLMs can also be multi-lingual.  For example, ChatGLM-6B and GLM-130B~\cite{zeng2023glm-130b} is a bilingual (English and Chinese) pretrained language model with 130 billion parameters.  LLMs can also be fine-tuned, a process by which the model parameters are adjusted with domain-specific data to align the performance of the LLM to a specific use case. For example, GPT-3 and GPT-4 \cite{openai2023gpt4} have been fine-tuned using reinforcement learning with human feedback (RLHF). 

\subsection{Vision Transformers} 
A Vision Transformer (ViT)~\cite{dosovitskiy2021an,han2022survey,khan2022transformers} is a transformer architecture for computer vision tasks including image classification segmentation, and object detection. A ViT treats an image as a sequence of image patches referred to as tokens. In the image tokenization process, an image is divided into patches of fixed size. Then the patches are flattened into a one-dimensional vector which is referred to as linear embedding. To capture the spatial relationships between image patches, positional information is added to each token. This process is referred to as position embedding. The image tokens incorporated with position encoding are fed into the transformer encoder and the self-attention mechanism enables the model to capture long-term dependencies and global patterns in the input data. In this paper, we focus only on those ViT models with a large number of parameters. ViT-G~\cite{zhai2022scaling} scales up the ViT model and has 2B parameters. Additionally, ViT-e~\cite{chen2022pali} has 4B parameters. ViT-22B~\cite{dehghani2023scaling} is a vision transformer model at 22 billion parameters, which is used in PaLM-E and PaLI-X~\cite{chen2023pali} and helps with robotics tasks.

DINO~\cite{caron2021emerging} is a self-supervised learning method, for training ViT. DINO is a form of knowledge distillation with no labels. Knowledge distillation is a learning framework where
a smaller model (student network) is trained to mimic the behavior of a larger more complex model (teacher network). Both networks share the same architecture with different sets of parameters. Given a fixed teacher network, the student network learns its parameters by minimizing the cross-entropy loss w.r.t. the student network parameters. The neural network architecture is composed of ViT or ResNet \cite{He2015} backbone and a projection head that includes layers of multi-layer perception (MLP). Self-supervised ViT features learned using DINO contain explicit information about the semantic segmentation of an image including scene layout and object boundaries with such clarity that is not achieved using supervised ViTs or convnets. 

DINOv2~\cite{oquab2023dinov2} provides a variety of pretrained visual models that are trained with different vision transformers (ViT) on the LVD-142M dataset introduced in~\cite{oquab2023dinov2}. It is trained using a discriminative self-supervised method on a compute cluster of 20 nodes equipped with 8 V100-32GB GPUs. DINOv2 provides various visual features at the image  (e.g. detection) or pixel level (e.g. segmentation).  SAM~\cite{kirillov2023segment} provides zero-shot promptable image segmentation. It is discussed in more detail in Section \ref{Perception}.   

\subsection{Multimodal Vision-Language Models (VLMs)}
Multimodal refers to the ability of a model to accept different ``modalities'' of inputs, for example, images, texts, or audio signals. Visual-language models (VLM) are a type of multi-modal model that takes in both images and text. A commonly used VLM in robotics applications is Contrastive Language-Image Pre-training (CLIP) \cite{CLIP2021}. CLIP offers a method to compare the similarity between textual descriptions and images. CLIP uses internet-scale image-text pairs data to capture the semantic information between images and text. CLIP model architecture contains a text encoder~\cite{radford2019language} and an image encoder (a modified version of vision transformer ViT) that are trained jointly to maximize the cosine similarity of the image and text embeddings. CLIP uses contrastive learning together with language models and visual feature encoders to incorporate models for zero-shot image classification.

BLIP~\cite{li2022blip} focuses on multimodal learning by jointly optimizing three objectives during pretraining. These objectives include Image-Text Contrastive Loss, Image-Text Matching Loss, and Language Modeling Loss. The method leverages noisy web data by bootstrapping captions, enhancing the training process.
CLIP$^2$~\cite{zeng2023clip2} 
aims to build well-aligned and instance-based text-image-point proxies. It learns semantic and instance-level aligned point cloud representations using a cross-modal contrastive objective.
FILIP~\cite{Yao2021FILIPFI} focuses on achieving finer-level alignment in multimodal learning. It incorporates a cross-modal late interaction mechanism that utilizes token-wise maximum similarity between visual and textual tokens. This mechanism guides the contrastive objective and improves the alignment between visual and textual information.
FLIP~\cite{li2023scaling} proposes a simple and more efficient training method for CLIP. FLIP randomly masks out and removes a significant portion of image patches during training. This approach aims to improve the training efficiency of CLIP while maintaining its performance.
 
\subsection{Embodied Multimodal Language Models}
An embodied agent is an AI system that interacts with a virtual or physical world. Examples include virtual assistance or robots. Embodied language models are foundation models that incorporate real-world sensor and actuation modalities into pretrained large language models.
Typical vision-language models are trained on general vision-language tasks such as image captioning or visual question answering. PaLM-E~\cite{driess2023palme} is a multimodal language model that has been trained on not only internet-scale general vision-language data, but also on embodied, robotics data, simultaneously.
In order to connect the model to real world sensor modalities, PaLM-E's architecture injects (continuous) inputs such as images, low-level states, or 3D neural scene representations into the language embedding space of a decoder-only language model to enable the model to reason about text and other modalities jointly.
The main PaLM-E version is built from the PaLM LLM~\cite{Chowdhery2022PaLMSL} and a ViT~\cite{dehghani2023scaling}. The ViT transforms an image into a sequence of embedding vectors which are projected into the language embedding space via an affine transformation.
The whole model is trained end-to-end, starting from a pre-trained LLM and ViT model.
The authors also explore different strategies such as freezing the LLM and just training the ViT, which leads to worse performance.
Given multimodal inputs, the output of PaLM-E is text decoded auto-regressively.
In order to connect this output to a robot for control, language conditioned short-horizon policies can be used. In this case, PaLM-E acts as a high-level control policy.
Experiments show that a single PaLM-E, in addition to being a vision-language generalist, is able to perform many different robotics tasks over multiple robot embodiments. 
The model exhibits positive transfer, i.e.\ simultaneously training on internet-scale language,  general vision-language, and embodied domains leads to higher performance compared to training the model on single tasks.

\subsection{Visual Generative Models}
Web-scale diffusion models such as OpenAI's DALL-E~\cite{pmlr-v139-ramesh21a} and DALL-E2 \cite{DALE2} provide zero-shot text-to-image generation. They are trained on hundreds of millions of image-caption pairs from the internet. These models learn a language-conditioned distribution over images from which an image can be generated using a given prompt. The DALL-E2 architecture includes a prior that generates a CLIP
image embedding from a text caption, and a decoder that generates an image conditioned on the image embedding. 

\begin{table*}[!ht]
\begin{threeparttable}
    \caption{Large Pretrained Models}
    \label{table:1}
    \begin{tabular} {C{0.1\textwidth}C{0.15\textwidth}C{0.1\textwidth}C{0.15\textwidth}C{0.12\textwidth}C{0.1\textwidth}C{0.1\textwidth}}
    \toprule
    Model   & Architecture & Size  & Training Data   & What to Pretrain & How to Pretrain  & Hardware \\
    \midrule
    CLIP \cite{CLIP2021} &  ViT-L/14@336px and a text encoder \cite{radford2019language}  & 0.307B & 400M image-text pairs & zero-shot image classification & contrastive pre-training & fine-tuned CLIP model is trained for 12 days on 256 V100
    GPUs \\
    \rowcolor{mygray}
    GPT-3 \cite{brown2020language}  & transformer (slight modification of GPT-2)& 175B  &Common Crawl (about a trillion words) & text output   & autoregressive model&  NPA \tnote{*}\\
    GPT-4 \cite{openai2023gpt4}&NPA & NPA  &  NPA& text output  & NPA & NPA \\
    \rowcolor{mygray}
    PaLI-X \cite{chen2023pali} & encoder-decoder & 55B & 10B image-text pairs from WebLI \cite{chen2022pali} and auxiliary tasks
    &  text and image to text output & autoregressive model & runs on multi-TPU cloud service\\
    DALL-E \cite{pmlr-v139-ramesh21a} & decoder-only transformer &  12B  & 250M text-image pairs  & zero-shot text-to-image generation & autoregressive model  & NPA \\
    \rowcolor{mygray}
    DALL-E2 \cite{DALE2}  &a  prior based on CLIP+ a decoder &  3.5B  & CLIP and DALL-E~\cite{pmlr-v139-ramesh21a}  & zero-shot text-to-image generation & diffusion  & NPA \\
    DINOv2 \cite{oquab2023dinov2} &  ViT-g/14 & 1.1B & LVD-142M \cite{oquab2023dinov2} & visual-features (image-level and pixel-level) & discriminative  & 20 nodes equipped with 8 V100-32GB GPUs  \\
    \rowcolor{mygray} 
    SAM \cite{kirillov2023segment}  & MAE \cite{he2022masked} vision transformer+CLIP \cite{radford2021learning} text encoder& 632M for ViT-H + 63M for CLIP text encoder
&SA-1B dataset~\cite{kirillov2023segment} that includes 1.1B segmentation masks on 11M images&   zero-shot promptable image segmentation  & supervised learning & 256 A100 GPUs for 68 hours\\
    \bottomrule
    \end{tabular}
    \begin{tablenotes}\footnotesize
\item[*] NPA stands for not publicly available.
\end{tablenotes}
\end{threeparttable}
\end{table*}

\section{Robotics}\label{Decision-Making}

In this section, we delve into robot decision-making, planning, and control. Within this realm, Large Language Models (LLMs) and Visual Language Models (VLMs) may hold the potential to serve as valuable tools for enhancing robotic capabilities. For instance, LLMs may facilitate the process of task specification, allowing robots to receive and interpret high-level instructions from humans. 
VLMs may also promise contributions to this field. VLMs specialize in the analysis of visual data. This visual understanding is a critical component of informed decision-making and complex task execution for robots. Robots can now leverage natural language cues to enhance their performance in tasks involving manipulation, navigation, and interaction. Vision-language goal-conditioned policy learning, whether through imitation learning or reinforcement learning, holds promise for improvement using foundation models. Language models also play a role in offering feedback for policy learning techniques. This feedback loop fosters continual improvement in robotic decision-making, as robots can refine their actions based on the feedback received from an LLM. This section underscores the potential contributions of LLMs and VLMs in robot decision-making. Assessing and comparing the contributions of papers in this section presents greater challenges compared to the other sections like the Perception Section (\ref{Perception}) or the Embodied AI Section (\ref{EmbodiedAI}). This is due to the fact that most papers in this section either rely on hardware experiments, using custom elements in the low-level control and planning stack that are not easily transferred to other hardware or other experimental setups, or they utilize non-physics-based simulators, which allow these low-level parts of the stack to be ignored, but leaving open the issue of non-transferability between different hardware implementations. In Section \ref{challenges and Future Directions}, we discuss the lack of benchmarking and reproducibility that needs to be addressed in future research. 

\subsection{Robot Policy Learning for Decision Making and Control}
In this section we discuss robot policy learning including language-conditioned imitation learning and language-assisted reinforcement learning.
\subsubsection{Language-conditioned Imitation Learning for Manipulation}

In language-conditioned imitation learning, a goal-conditioned policy $\pi_{\theta}(a_t | s_t, l)$ is learned that outputs actions $a_t \in \mathcal{A}$ conditioned on the current state $s_t \in \mathcal{S}$ and language instruction $l \in \mathcal{L}$.
The loss function is defined as the maximum likelihood goal conditioned imitation objective:
\begin{equation}\label{eq:BC}
   \mathcal{L}_{\textrm{GCIL}} = \mathbb{E}_{(\tau, l) \sim \mathcal{D}} \sum_{t=0}^{|\tau|} \textrm{log}\pi_{\theta}(a_t | s_t, l),
\end{equation}
where $\mathcal{D}$ is the language-annotated demonstration dataset $\mathcal{D} = \{\tau_i\}_i^N$. Demonstrations can be represented as trajectories, or sequences of images, RGB-D voxel observations, etc. Language instructions are paired with demonstrations to be used as the training dataset. Each language-annotated demonstration $\tau_i$ consists of $\tau_i = \{(s_1, l_1, a_1), (s_2, l_2, a_2), ...\}$.  
 At test time, the robot is given a series of instructions and the language-conditioned visuomotor policy $\pi_{\theta}$ provides actions $a_t$ in a closed loop given the instruction at each time step.  The main challenges in this domain are: (i) obtaining a sufficient volume of demonstrations and conditioning labels to train a policy, (ii) distribution shift under the closed-loop policy---the feedback of the policy can lead the robot into regions of the state space that are not well-covered in the training data, negatively impacting performance. (All the following papers in this subsection focus on robot manipulation tasks.)
 
Since generating language-annotated data by pairing demonstrations with language instruction is an expensive process, the authors in Play-LMP \cite{lynch2019play} propose learning from teleoperated play data. In this setting, reusable latent plan representations are learned from unlabeled play data. Also, a goal-conditioned policy is learned to decode the inferred plan to perform the task specified by the user. 
In addition, the distributional shift in imitation learning is analyzed and it is shown in this setting that the play data is more robust with respect to perturbation compared to expert positive demonstrations. Note that language goal $l$ in (\ref{eq:BC}) can be substituted with any other type of goal for example goal image, which is another common choice of goal in goal-conditioned imitation learning.   

In a follow-up work~\cite{lynch2021language}, the authors present multi-context imitation (MCIL) which uses language-conditioned imitation learning over unstructured data. The multi-Context imitation framework is based on relabeled imitation learning and labeled instruction following. MCIL assumes access to multiple contextual imitation datasets, for example, goal image demonstrations, language goal demonstrations, or one-hot task demonstrations. MCIL trains a single latent goal-conditioned policy over all datasets simultaneously by encoding contexts in the shared latent space using the associated encoder for each context. Then a goal-conditioned imitation loss is computed by averaging over all datasets. The policy and goal-encoders are trained end-to-end. Another approach to tackle the data annotation challenge in language-conditioned imitation learning involves utilizing foundation models to offer feedback by labeling demonstrations. In~\cite{PAFF2023}, the authors propose to use pretrained foundation models to provide feedback. To deploy a trained policy to a new task or new environment, the policy is played using randomly generated instructions, and a pretrained foundation model provides feedback by labeling the demonstration. Also, this paired instruction-demonstration data can be used for policy fine-tuning. CLIPort~\cite{shridhar2021cliport} also presents a language-conditioned imitation learning for vision-based manipulation. A two-stream architecture is presented that combines the semantic understanding of CLIP with the spatial precision of Transporter~\cite{zeng2020transporter}. This end-to-end framework solves language-specified manipulation tasks without any explicit representation of the object poses or instance segmentation. CLIPort grounds semantic concepts in precise spatial reasoning, but it is limited to 2D observation and action spaces. To address this limitation, the authors of PerAct (Perceiver-Actor)~\cite{shridhar2022peract} propose to represent observation and action spaces with 3D voxels and employ the 3D structure of voxel patches for efficient language-conditioned behavioral cloning with transformers to imitate 6-DoF manipulation tasks from just a few demonstrations. While 2D behavioral cloning methods such as CLIPort are limited to single-view observations, 3D approaches such as PerAct allow for multi-view observations as well as 6-DoF action spaces. PerAct uses only CLIP's language encoder to encode the language goal. PerAct takes language goals and RGB-D voxel observations as inputs to a Perceiver Transformer and outputs discretized actions by detecting the next best voxel action. PerAct is trained through supervised learning with discrete-time input actions from the demonstration dataset. The demonstration dataset includes voxel observations paired with language goals and keyframe action sequences. An action consists of a 6-DoF pose, gripper open state, and collision avoidance action. During training, a tuple is randomly sampled and the agent predicts the keyframe action given the observation and goal.

Grounding semantic representations into a spatial environment is essential for effective robot interaction. CLIPort and PerAct utilize CLIP (which is trained based on contrastive learning) for semantic reasoning and Transporter and Perceiver for spatial reasoning.   

Voltron~\cite{karamcheti2023language} presents a framework for language-driven representation learning in robotics. Voltron captures semantic, spatial, and temporal representations that are learned from videos and captions.  Contrastive learning captures semantic representations but loses spatial relationships, and in contrast, masked autoencoding captures spatial and not semantic representations. Voltron trades off language-conditioned visual reconstruction for local spatial representations and visually-grounded language generation to capture semantic representations. This framework includes grasp affordance prediction, single-task visuomotor control, referring expression grounding, language-conditioned imitation, and intent-scoring tasks. Voltron models take videos and their associated language captions as input to a multimodal encoder whose outputs are then decoded to reconstruct one or more frames from a masked context. Voltron starts with a masked autoencoding backbone and adds a dynamic component to the model by conditioning the MAE encoder on a language prefix. Temporal information is captured by conditioning on multiple frames.

Deploying robot policy learning techniques that leverage language-conditioned imitation learning with real robots presents ongoing challenges. These models rely on end-to-end learning, where the policy maps pixels or voxels to actions. As they are trained through supervised learning on demonstration datasets, they are susceptible to issues related to generalization and distribution shifts. To improve robustness and adaptability, techniques such as data augmentation and domain adaptation can make the policies more robust to the distribution shift. 

CACTI~\cite{mandi2022cacti} is a novel framework designed to enhance scalability in robot learning using foundation models such as Stable Diffusion~\cite{LatentDiffusion2022}. CACTI introduces the four stages of data collection, data augmentation, visual representation learning, and imitation policy training. In the data collection stage, limited in-domain expert demonstration data is collected. In the data augmentation stage, CACTI employs visual generative models such as Stable Diffusion~\cite{LatentDiffusion2022} to boost visual diversity by augmenting the data with scene and layout variations. In the visual representation learning stage, CACTI leverages pretrained zero-shot visual representation models trained on out-of-domain data to improve training efficiency. Finally, in the imitation policy training stage, a general multi-task policy is learned using imitation learning on the augmented dataset with compressed visual representations as input. CACTI is trained for multi-task and multi-scene manipulation in kitchen environments, both in simulation and the real world. The use of these techniques enhances the generalization ability of the framework and enables it to learn from a wide range of environments.

Beyond language, recent works have investigated other forms of task specification. Notably, MimicPlay \cite{wang2023mimicplay} presents a hierarchical imitation learning algorithm that learns high-level plans in latent spaces from human play data and low-level motor commands from a small number of teleoperated demonstrations. By harnessing the complementary strengths of these two data sources, this algorithm can significantly reduce the cost of training visuomotor policies for long-horizon manipulation tasks. Once trained, it is capable of performing new tasks based on one human video demonstration at test time. MUTEX~\cite{shah2023mutex} further explores learning a unified policy across multimodal task specifications in video, image, text, and audio, showing improved policy performances over single-modality baselines through cross-modal learning.

\subsubsection{Language-Assisted Reinforcement Learning}

Reinforcement learning (RL) is a family of methods that enable a robot to optimize a policy through interaction with its environment by optimizing a reward function. These interactions are usually in a simulation environment, sometimes augmented with data from physical robot hardware for sim-to-real transfer.  RL has close ties to optimal control.  Unlike imitation learning, RL does not require human demonstrations, and (in theory) has the potential to attain super-human performance. In the RL problem, the expected return of a policy is maximized using the collected roll-outs from interactions with the environment. The feedback received from the environment in the form of a reward signal guides the robot to learn which actions lead to favorable results and which do not. In this section, we discuss works that have incorporated foundation models (LLM, VLMs, etc.) into RL problems. 

Fast and flexible adaptation is a desired capability of artificial agents and is essential for progress toward general intelligence. In Adaptive Agent (AdA)~\cite{adaptiveagentteam2023humantimescale} the authors present an RL foundation model that is an agent pretrained on diverse tasks and is designed to quickly adapt to open-ended embodied 3D problems by using fast in-context learning from feedback. This work considers navigation, coordination, and division of labor tasks. Given a few episodes within an unseen environment at test time, the agent engages in trial-and-error exploration to refine its policy toward optimal performance. In AdA a transformer architecture is trained using model-based RL$^2$ \cite{duan2017rl} to train agents with large-scale attention-based memory, which is required for adaptation. Transformer-XL \cite{dai2019transformerxl} with some modification is used to enable long and variable-length context windows to increase the model memory to capture long-term dependencies. The agent collects diverse data in the XLand environment that includes $10^{40}$ possible tasks \cite{deepmind_open_ended_learning2021}, in an automated curriculum. In addition, distillation is used to enable scaling to models with more than 500M parameters.  

Palo et al.~\cite{palo2023towards} propose an approach to enhance reinforcement learning by integrating Large Language Models (LLMs) and Visual-Language Models (VLMs) to create a more unified RL framework. This work considers robot manipulation tasks. Their approach addresses core RL challenges related to exploration, experience reuse and transfer, skills scheduling, and learning from observation. The authors use an LLM to decompose complex tasks into simpler sub-tasks, which are then utilized as inputs for a transformer-based agent to interact with the environment. The agent is trained using a combination of supervised and reinforcement learning, enabling it to predict the optimal sub-task to execute based on the current state of the environment.

\subsection{Language-Image Goal-Conditioned Value Learning} 
In value learning, the aim is to construct a value function that aligns goals in different modalities and preserves temporal coherence due to the recursive nature of the value function. Reusable Representation for Robotic Manipulations (R3M)~\cite{nair2022r3m} provides pretrained visual representation for robot manipulation using diverse human video datasets such as Ego4D and can be used as a frozen perception module for policy learning in robot manipulation tasks. R3M's pretrained visual representation is demonstrated on Franka Emika Panda's arm and enables different downstream manipulation tasks. R3M is trained using time-contrastive learning to capture temporal dependencies, video-language alignment to capture semantic features of the scene (such as objects and their relationships) and $L1$ penalty to encourage sparse and compact representation. For a batch of videos, using time-contrastive loss, an encoder is trained to generate a representation wherein the distance between images that are temporally closer is minimized compared to images that are farther apart in time or from different videos. 

Similar to R3M, Value-Implicit Pretraining (VIP)~\cite{VIP2022} employs time-contrastive learning to capture temporal dependencies in videos, but it does not require video-language alignment. VIP is also focused on robot manipulation tasks. VIP is a self-supervised approach for learning visual goal-conditioned value functions and representations from videos. VIP learns visual goal-based rewards for downstream tasks and can be used for zero-shot reward specification. The reward model is derived from pretrained visual representations. Pretraining involves using unlabeled human videos. Human videos do not contain any action information to be used for robot policy learning, therefore the learned value function does not explicitly depend on actions. VIP introduces a novel time contrastive objective that generates a temporally smooth embedding. The value function is implicitly defined via distance embedding. The proposed implicit time contrastive learning attracts the representation of the initial and goal frames in the same trajectory and repels the representation of intermediate frames by recursive one-step temporal difference minimization. This representation captures long-term temporal dependencies across task frames and local temporal smoothness among adjacent frames. 

Language-Image Value Learning (LIV)~\cite{ma2023liv} is a control-centric vision-language representation. LIV generalizes the prior work VIP by learning multi-modal vision-language value functions and representations using language-aligned videos. 
For tasks specified as language goals or image goals, a multi-model representation is trained that encodes a universal value function. LIV is also focused on robot manipulation tasks. LIV is a pretrained control-centric vision-language representation based on large human video datasets such as EPIC-KITCHENS~\cite{Damen2018EPICKITCHENS}.
The representations are kept frozen during policy learning. A simple MLP is used on top of pretrained representations for the policy network. Policy learning is decoupled from language-visual representation pretraining. The LIV model is pretrained on arbitrary video activity datasets with text annotation, and the model can be fine-tuned on small datasets of in-domain robot data to ground language in a context-specific way. LIV uses a generalization of the mutual information-based image-text contrastive representation learning objective as used in CLIP, so LIV can be considered as a combination of CLIP and VIP. 
Both VIP and LIV learn a self-supervised goal-conditioned value-function objective using contrastive learning. The LIV extends the VIP framework to multi-modal goal specifications. LOREL~\cite{nair2022a} learns a language-conditioned reward from offline data and uses it during model predictive control to complete language-specified tasks.

Value functions can be used to help ground semantic information obtained from an LLM to the physical environment in which a robot is operating. By leveraging value functions, a robot can associate the information processed by the LLM with specific locations and objects in its surroundings. In SayCan~\cite{Ahn2022DoAI}, researchers investigate the integration of large language models with the physical world through learning. They use the language model to provide task-grounding (Say), enabling the determination of useful sub-goals based on high-level instructions, and a learned affordance function to achieve world-grounding (Can), enabling the identification of feasible actions to execute the plan. 
Inner Monologue~\cite{huang2022inner} studies the role of grounded environment feedback provided to the LLM, thus closing the loop with the environment. The feedback is used for robot planning with large language models by leveraging a collection of perception models (e.g., scene descriptors and success detectors) in tandem with pretrained language-conditioned robot skills. Feedback includes task-specific feedback, such as success detection, and scene-specific feedback (either ``passive'' or ``active''). In both SayCan and Inner Monologue robot manipulation and navigation tasks are considered using a real-world mobile manipulator robot from Everyday Robots. Text2Motion~\cite{lin2023text2motion} is a language-based planning framework for long-horizon robot manipulation. Similar to SayCan and Inner Monologue, Text2Motion computes a score ($S_{\textrm{LMM}}$) associated with each skill at each time step. The task planning problem is to find a sequence of skills by maximizing the likelihood of a skill sequence given a language instruction and the initial state. In Text2Motion, the authors propose to verify that the generated long-horizon plans are symbolically correct and geometrically feasible. Hence, a geometric feasibility score ($S_{geo}$) is defined as the probability that all the skills in the sequence achieve rewards. To compute the overall score, the LLM score is multiplied by the geometric feasibility score ($S_{\textrm{Skill}} = S_{\textrm{LMM}}\cdot S_{\textrm{geo}}$).       

VoxPoser~\cite{VoxPoser2023} builds 3D value maps to ground affordances and constraints into the perceptual space. VoxPser considers robot manipulation tasks. Given the RGB-D observation of the environment and language instruction, VoxPoser utilizes large language models to generate code, which interacts with vision-language models to extract a sequence of 3D affordance maps and constraint maps. These maps are composed together to create 3D value maps. The value maps are then utilized as objective functions to guide motion planners to synthesize trajectories for everyday manipulation tasks without requiring any prior training or instruction. 

In~\cite{mahmoudieh2022zeroshot},
reward shaping using CLIP is presented. This work considers robot manipulation tasks. The proposed model utilizes CLIP to ground objects in a scene described by the goal text paired with spatial relationship rules to shape the reward by using raw pixels as input. They use developments in building large-scale visuo-lingual models like CLIP to devise a framework that generates the task reward signal from just the goal text description and raw pixel observations. This signal is then used to learn the task policy.

In  \cite{mees23hulc2}, Hierarchical Universal Language Conditioned Policies 2.0 (HULC++) is presented. This work considers robot manipulation tasks. A self-supervised visuo-lingual affordance model is used to learn general-purposed language-conditioned robot skills from unstructured offline data in the real world. This method requires annotating as little as 1\% of the total data with language. The visuo-lingual affordance model has an encoder-decoder architecture with two decoder heads. Both heads share the same encoder and are conditioned on the input language instruction. One head predicts a distribution over the image, in which each pixel likelihood is an afforded point. The other head predicts a Gaussian distribution from which the corresponding predicted depth is sampled. Given visual observations and language instructions as input, the affordance model outputs a pixel-wise heat map that represents affordance regions and the corresponding depth map. 

\subsection{Robot Task Planning using Large Language Models}
LLMs can be used to provide high-level task planning for performing complex long-horizon robot tasks.  
\subsubsection{Language Instructions for Task Specification}
As discussed above, SayCan~\cite{Ahn2022DoAI} uses an LLM for high-level task planning in language, though with a learned value function to ground these instructions in the environment.

Temporal logic is useful for imposing temporal specifications in robotic systems. 
In~\cite{chen2023nl2tl}, translation from natural language (NL) to temporal logic (TL) is proposed. A dataset with 28k NL-TL pairs is created and the T5 \cite{Raffel2020}
model is finetuned using the dataset. LLMs are often used to plan task sub-goals. This work considers robot navigation tasks. In \cite{Chen2023}, instead of direct task planning, a few-shot translation from a natural language task description to an intermediary task representation is performed. This representation is used by a Task and Motion Planning (TAMP) algorithm to jointly optimize task and motion plans. Autoregressive re-prompting is used to correct synthetic and semantic errors. This work also considers robot navigation tasks.  

\subsubsection{Code Generation using Language Models for Task Planning}
Classical task planning requires extensive domain knowledge and the search space is large~\cite{huang2021learning,sun2021neuro}. 
LLMs can be used to generate sequences of tasks required to achieve a high-level task. In ProgPrompt~\cite{progprompt2023}, the authors introduce a prompting method that uses LLMs to generate sequences of actions directly with no additional domain knowledge. The prompt to the LLM includes specifications of the available actions, objects in the environment, and example programs that can be executed. VirtualHome~\cite{puig2018virtualhome} is used as a simulator for demonstration. 

Code-as-Policies~\cite{codeaspolicies2022} explores the use of code-writing LLMs to generate robot policy code based on natural language commands. This work considers robot manipulation and navigation tasks using a real-world mobile manipulator robot
from Everyday Robots. The study demonstrates that LLMs can be repurposed to write policy code by expressing functions or feedback loops that process perception outputs and invoke control primitive APIs.
To achieve this, the authors utilize few-shot prompting, where example language commands formatted as comments are provided along with the corresponding policy code. Without any additional training on this data, they enable the models to autonomously compose API calls and generate new policy code when given new commands.
The approach leverages classic logic structures and references third-party libraries like NumPy and Shapely to perform arithmetic operations. By chaining these structures and using contextual information (behavioral commonsense), the LLMs can generate robot policies that exhibit spatial-geometric reasoning, generalize to new instructions, and provide precise values (e.g., velocities) for ambiguous descriptions such as ``faster.''
The concept of ``code as policies'' formalizes the generation of robot policies using language model-generated programs (LMPs). These policies can represent reactive policies like impedance controllers, as well as waypoint-based policies such as vision-based pick and place or trajectory-based control. The effectiveness of this approach is demonstrated on multiple real robot platforms.
A crucial aspect of this approach is the hierarchical code generation process, which involves recursively defining undefined functions. This enables the LLMs to generate more complex code structures to fulfill the desired policy requirements. 

In \cite{vemprala2023chatgpt}, the authors provide design principles for using ChatGPT in robotics and demonstrate how LLMs can help robotic capabilities rapidly generalize to different form factors. This work considers robot manipulation and aerial navigation tasks. First, a high-level robot function library that maps to multiple atomic tasks executable by the robot is defined. Then, a prompt is crafted that includes these functions, and the required constraints along the task description. ChatGPT then provides executable
code specific to the given robot configuration and task. The generated code can then be evaluated by a user and appropriate feedback with modified prompts to LLMs further help refine and generate programs that are safe and deployable on the physical robot. The study demonstrated that such a methodology can be applied to multiple form factors both in simulation and in the real world.

\subsection{In-context Learning (ICL) for Decision-Making} 

In-context Learning (ICL)\cite{dong2022survey} operates without the need for parameter optimization, relying instead on a set of examples included in the prompt (the concept of prompting). This learning approach is intimately linked with prompt engineering and finds extensive use in natural language processing. The method of Chain-of-Thought\cite{wei2022chain} is a prominent technique within in-context learning. It involves executing a sequence of intermediate steps to arrive at the final solution for complex, multi-step problems. This technique allows models to produce step-by-step explanations that parallel human cognitive processes. However, despite its numerous benefits, ICL also faces certain challenges, including issues related to ambiguity and interpretation, domain-specific knowledge, transparency, and explainability.
In-context learning has had a significant impact on the field of LLMs in a broad sense, and many robotics works have used it to apply LLMs to specific domains. 
Investigating this, Mirchandani and colleagues~\cite{generalpatternmachines2023} illustrate that Large Language Models (LLMs) possess remarkable pattern recognition abilities. They reveal that, through in-context learning, LLMs can effectively handle general patterns that extend beyond standard language-based prompts. This capability allows for the application of LLMs in scenarios such as offline trajectory optimization and online, in-context reinforcement learning.
Additionally, Jia and the team in their work on Chain-of-Thought Predictive Control~\cite{jia2023chain} suggest a method to identify specific brief sequences within demonstrations, termed as 'chain-of-thought'. They focus on understanding and representing the hierarchical structure of these sequences, highlighting the achievement of subgoals within tasks. This work considers robot policy learning from demonstrations for contact-rich object manipulation tasks. 

\subsection{Robot Transformers}
Foundation models can be used for end-to-end control of robots by providing an integrated framework that combines perception, decision-making, and action generation. 

Xiao et al.~\cite{Xiao2022} demonstrate the effectiveness of self-supervised visual pretraining using real-world images for learning motor control tasks directly from pixel inputs. This work is focused on robot manipulation tasks. They show that without any task-specific fine-tuning of the pretrained encoder, the visual representations can be utilized for various motor control tasks. This approach highlights the potential of leveraging self-supervised learning from real-world images to acquire general visual representations that can be applied across different motor control tasks.
Similarly, Radosavovic et al.~\cite{Radosavovic2022} investigate the use of self-supervised visual pretraining on diverse, in-the-wild videos for real-world robotic tasks. This work considers robot manipulation tasks. They find that the pretrained representations obtained from such videos are effective in a range of real-world robotic tasks, considering different robotic embodiments. This suggests that the learned visual representations generalize well across various tasks and robot platforms, demonstrating the broad applicability of self-supervised pretraining for real-world robotic applications.
Both studies emphasize the advantages of self-supervised visual pretraining, where models are trained on large amounts of unlabeled data to learn useful visual representations. By leveraging real-world images and videos, these approaches enable learning from diverse and unstructured visual data, leading to more robust and transferable representations for motor control tasks in robotic systems.

Another example of a Transformer-based policy model is the work on Robotics Transformer (RT-1)~\cite{RT12023}, where the authors demonstrate a model that shows promising scalability properties. To train the model, the authors use a large dataset of over 130k real-world robotic experiences, comprising more than 700 tasks, that was collected over 17 months using a fleet of 13 robots. RT-1 receives images and natural language instructions as inputs and outputs discretized base and arm actions. It can generalize to new tasks, maintain robustness in changing environments, and execute long-horizon instructions. The authors also demonstrate the model's capability to effectively absorb data from diverse domains, including simulations and different robots. 

The follow-up work, called Robotic Transformer 2 (RT-2)~\cite{rt22023arxiv}, demonstrates a vision-language-action (VLA) model that takes a step further by learning from both web and robotics data. The model effectively utilizes this data to generate generalized actions for robotic control. To do so, the authors use pre-existing vision-language models and directly co-fine-tune them on robot trajectories resulting in a single model that operates as a language model, a vision-language model, and a robot policy. To make co-fine-tuning possible, the actions are represented as simple text strings which are then tokenized using an LLM tokenizer into text tokens. The resulting model, RT-2, enables vision-language models to output low-level closed-loop control. Similarly to RT-1, actions are produced based on robot instructions paired with camera observations and the action space includes 6-DoF positional and rotational displacement of the robot end-effector, gripper extension, and episode termination command. Via extensive experiments, the authors show that utilizing VLMs aids in the enhancement of generalization across visual and semantic concepts and enables the robots to respond to the so-called chain of thought prompting, where the agent performs more complex, multi-stage semantic reasoning. Both RT-1 and RT-2 consider robot manipulation and navigation tasks using a real-world mobile manipulator robot
from Everyday Robots. 
One key limitation of RT-2 and other related works in robotics is the fact that the range of physical skills exhibited by the robot is limited to the distribution of skills observed within the robot's data. While one way to approach this limitation is to collect more diverse and dexterous robotic data, there might be other intriguing research directions such as using motion data in human videos, robotic simulations, or other robotic embodiments.

The next work utilizing the Transformer architecture indeed focuses on learning from data that combines multiple robotic embodiments. In RT-X~\cite{padalkar2023rtx}, the authors provide a number of datasets in a standardized data format and models to make it possible to explore the possibility of training large cross-embodied robotic models in the context of robotic manipulation. In particular, they assembled a dataset from 22 different robots collected through a collaboration between 21 institutions, demonstrating 527 skills (160266 tasks). 
With this unified dataset, RT-X demonstrates that RT-1- and RT-2-based models trained on this multi-embodiment, diverse data exhibit positive transfer across robotic domains and improve the capabilities of multiple robots by leveraging experience from other platforms.

Other works have investigated general pretrained transformers for robot control, trained with self-supervised trajectory data from multiple robots.  For example, Perception-Action Causal Transformer (PACT) \cite{Bonatti2023} is a generative transformer architecture that builds representations from robot data with self-supervision. This work considers robot navigation tasks. PACT pretrains a representation useful for multiple tasks on a given robot. Similar to how large language models learn from extensive text data, PACT is trained on abundant safe state-action data (trajectories) from a robot, learning to predict appropriate safe actions.
By predicting states and actions over time in an autoregressive manner, the model implicitly captures dynamics and behaviors specific to a robot. PACT was tested in experiments involving mobile agents: a wheeled robot with a LiDAR sensor (MuSHR) and a simulated agent using first-person RGB images (Habitat). The results show that this robot-specific representation can serve as a starting point for tasks like safe navigation, localization, and mapping.
Additionally, the experiments demonstrated that fine-tuning smaller task-specific networks on the pre-trained model leads to significantly better performance compared to training a single model from scratch for all tasks simultaneously, and comparable performance to training a separate large model for each task independently.

Another work in this space is Self-supervised Multi-task pretrAining with contRol Transformer (SMART)~\cite{sun2023smart}, which introduces a self-supervised multi-task pertaining to control transformers, providing a pretraining-finetuning approach tailored for sequential decision-making tasks. During the pretraining phase, SMART captures information essential for both short-term and long-term control, facilitating transferability across various tasks. Subsequently, the finetuning process can adapt to a wide variety of tasks spanning diverse domains. Experimentation underscores SMART's ability to enhance learning efficiency across tasks and domains. This work considers cart pole-swing-up, cart pole-balance, hopper-hop, hopper-stand,
cheetah-run, walker-stand walker-run, and walker-walk tasks. The approach demonstrates robustness against distribution shifts and proves effective with low-quality pretraining datasets.

Some works have investigated transformer models in conjunction with classical planning and control layers as part of a modular robot control architecture.  For example, in \cite{Bucker2023}, a multi-modal transformer (LATTE) is presented that allows a user to reshape robot trajectories using language instructions. This work considers both robot manipulation and navigation tasks. LATTE transformer takes as input geometrical features of an initial trajectory guess along with the obstacle map configuration, language instructions from a user, and images of each object in the environment. The model's output is modified for each waypoint in the trajectory so that the final robot motion can adhere to the user's language instructions. The initial trajectory plan can be generated using any geometric planner such as A$^\ast$, RRT$^\ast$, or model predictive control. Subsequently, this plan is enriched with the semantic objectives within the model. LATTE leverages pretrained language and visual-language models to harness semantic representations of the world.

\subsection{Open-Vocabulary Robot Navigation and Manipulation}\label{Applications in Robotics}
\subsubsection{Open-Vocabulary Navigation} Open-vocabulary navigation addresses the challenge of navigating through unseen environments. The open-vocabulary capability signifies that the robot possesses the capacity to comprehend and respond to language cues, instructions, or semantic information, without being restricted to a predefined dataset. In this section, we explore papers that examine the integration of LLMs, VLMs, or a combination of both in a plug-and-play manner for robot navigation tasks. Additionally, we discuss papers that take a different approach by constructing foundation models explicitly tailored for robot navigation tasks.

In VLN-BERT~\cite{majumdar2020improving}, the authors present a visual-linguistic transformer-based model that leverages multi-modal visual and language representations for visual navigation using web data. The model is designed to score the compatibility between an instruction, such as ``...stop at the brown sofa," and a sequence of panoramic RGB images captured by the agent.

Similarly, LM-Nav~\cite{shah2022lmnav} considers visual navigation tasks. LM-Nav is a system that utilizes pretrained models of images and language to provide a textual interface to visual navigation. LM-Nav demonstrates visual navigation in a real-world outdoor environment from natural language instructions. LM-Nav utilizes an LLM (GPT-3~\cite{brown2020language}), a VLM (CLIP~\cite{CLIP2021}), and a VNM (Visual Navigation Model). First, LM-Nav constructs a topological graph of the environment via the VNM estimating the distance between images. The LLM is then used to translate the natural instructions to sequences of intermediate language landmarks. The VLM is used to ground the visual observations in landmark descriptions via a joint probability distribution over landmarks and images. Using the VLM's probability distribution, the LLM instructions, and the VNM's graph connectivity, the optimal path is planned using the search algorithm. Then the plan is executed by the goal-conditioned policy of VNM. 

While LM-Nav makes use of LLMs and VLMs as plug-and-play for visual navigation tasks, the authors of ViNT~\cite{shah2023vint} propose to build a foundation model for visual navigation tasks. ViNT is an image goal-conditioned navigation policy trained on diverse training data and can control different robots in zero-shot. It can be fine-tuned to be adapted for different robotic platforms and various downstream tasks. ViNT is trained on various navigation datasets from different robotic platforms. It is trained with goal-reaching objectives and utilizes a Transformer-based architecture to learn navigational affordances. ViNT encodes visual observations and visual goals using an EfficientNet CNN and predicts temporal distance and normalized actions in an embodiment-agnostic manner. Additionally, ViNT can be augmented with diffusion-based sub-goal proposals to help explore environments not encountered during training. An image-to-image diffusion generates sub-goal images, which the ViNT then navigates toward while building a topological map in the background.

Another work that considers zero-shot navigation tasks is Audio Visual Language Maps (AVLMaps)~\cite{huang23avlmaps}. AVLMaps presents a 3D spatial map representation for cross-modal information from audio, visual, and language cues. AVLMaps receives multi-modal prompts and performs zero-shot navigation tasks in the real world. The inputs are depth and RGB images, camera pose, and audio. Visual features are encoded using pretrained foundation models. Visual localization features (using NetVLAD~\cite{Arandjelovic16}, SuperPoint~\cite{detone18superpoint}), visual-language features (using LSeg~\cite{li2022languagedriven}), and audio-language features (using AudioCLIP~\cite{guzhov2021audioclip}) are computed and predictions from different modalities are combined into 3D heatmaps. The pixel-wise joint probability of the heatmap is computed and used for planning. Additionally, navigation policies are generated as executable codes with the help of GPT-3. 
Finally, 3D heatmaps are predicted indicating the location of multimodal concepts such as objects, sounds, and images.
 
Many roboticists may wonder about the comparative strengths of classical modular robot navigation systems versus end-to-end learned systems. 
Semantic navigation~\cite{semanticNav2023} seeks to address this question by presenting an empirical analysis of semantic visual navigation methods. The study compares representative approaches from classical, modular, and end-to-end learning paradigms across six different homes, without any prior knowledge, maps, or instrumentation.
The findings of the study reveal that modular learning methods perform well in real-world scenarios. In contrast, the end-to-end learning approaches face challenges due to a significant domain gap between simulated and real-world images. This domain gap hinders the effectiveness of end-to-end learning methods in real-world navigation tasks.
For practitioners, the study emphasizes that modular learning is a reliable approach to object navigation. The modularity and abstraction in policy design enable successful transfer from simulation to reality, making modular learning an effective choice for practical implementations.
For researchers, the study also highlights two critical issues that limit the reliability of current simulators as evaluation benchmarks. Firstly, there exists a substantial Sim-to-Real gap in images, which hampers the transferability of learned policies from simulation to the real world. Secondly, there is a disconnect between simulation and real-world error modes, which further complicates the evaluation process.

Another line of work in open-vocabulary navigation is object navigation tasks. In this task, the robot must be able to find the object described by humans and navigate towards the object. The navigation task is decomposed into exploration when the language target is not detected and exploitation when the target is detected and the robot navigates toward the target. As the robot moves in the environment, it creates a top-down map using RGB-D observations and poses estimates. In~\cite{gadre2022cow}, the authors introduce a zero-shot object navigation setting that uses an open-vocabulary classifier such as CLIP~\cite{CLIP2021} to compute the cosine similarity between an image and a user-specified description.     

Common datasets and benchmarks for these types of problems are Matterport3D~\cite{Matterport3D,mattersim}, Gibson~\cite{li2021igibson} and Habitat~\cite{szot2021habitat}.
L3MVN~\cite{Yu2023} enhances visual target navigation by constructing an environment map and selecting long-term goals using the inference capabilities of large language models. The system can determine appropriate long-term goals for navigation by leveraging pretrained language models such as RoBERTa-large~\cite{liu2019roberta}, enabling efficient exploration and searching. Chen et al.~\cite{chen2023train} presents a training-free and modular system for object goal navigation, which constructs a structured scene representation through active exploration. The system utilizes semantic information in the scene graphs to deduce the location of the target object and integrates semantics with the geometric frontiers to enable the agent to navigate effectively to the most promising areas for object search while avoiding detours in unfamiliar environments.
HomeRobot~\cite{homerobot} introduces a benchmark for the Open-Vocabulary Mobile Manipulation (OVMM) task. OVMM task is the problem of finding an object in any unseen environment, navigating towards the object, picking it up, and navigating towards a goal location to place the object. HomeRobot provides a benchmark in simulation and the real world for OVMM tasks.
       
\subsubsection{Open-Vocabulary Manipulation}
Open-vocabulary manipulation refers to the problem of manipulating any object in a previously unseen environment. 
VisuoMotor Attention Agent (VIMA)~\cite{jiang2023vima} learns robot manipulation from multi-modal prompts. VIMA is a transformer-based agent that predicts motor commands conditioned on a task prompt and a history of interactions. VIMA  It introduces a new form of task specifications that combines textual and visual tokens. Multi-modal prompting converts different robot manipulation tasks, such as visual goal-reaching, learning from visual demonstrations, and novel concept grounding into one sequence modeling problem. It offers the training of a unified policy across diverse tasks, potentially allowing for zero-shot generalization to previously unseen ones. VIMA-BENCH is introduced as a benchmark for multi-modal robot learning. The VIMA-BENCH simulator supports collections of objects and textures that can be utilized in multi-modal prompting. RoboCat~\cite{Bousmalis2023RoboCat} is a self-improving AI agent. It uses a 1.18B-parameter decoder-only transformer. It learns to operate different robotic arms, solves tasks from as few as 100 demonstrations, and improves from self-generated data. RoboCat is based on Gato~\cite{reed2022generalist} architecture and is trained with a self-improvement cycle.

For robots to operate effectively in the real world they must be able to manipulate previously unseen objects. Liu et al. present StructDiffusion~\cite{liu2022structdiffusion}, which seeks to enable robots to use partial viewpoint clouds and natural language instructions to construct a goal configuration for objects that were previously seen or unseen. They accomplish this by first using segmentation to break up the scene into objects. Then they use a multi-model transformer to combine word and point cloud embeddings and output a 6-DoF goal pose prediction. The predictions are iteratively refined via diffusion and a discriminator that is trained to determine if a sampled configuration is feasible. Manipulation of Open-World Objects (MOO)~\cite{moo2023arxiv} leverages a pretrained vision-language model to extract object-centric information from the language command and the image and conditions the robot policy on the current image, the instructions, and the extracted object information in a form of a single-pixel overlaid onto the image. MOO uses Owl-ViT for object detection and RT-1 for language-conditioned policy learning. 

Another task in robot manipulation involves autonomous scene rearrangement and in-painting. 
DALL-E-Bot~\cite{dall_e_bot2023} performs zero-shot autonomous rearrangement in the scene in a human-like way using pretrained image diffusion model DALL-E2~\cite{DALE2}. DALL-E-Bot autonomous object rearrangement does not require any further data collection or training. First, the initial observation image (of the disorganized scene) is converted into a per-object representation including a segmentation mask using Mask R-CNN~\cite{MaskRCNN2017}, an object caption, and a CLIP visual feature vector. Then a text prompt is generated by describing the object in the scene and is given to DALL-E to create a goal image for the rearrangement task (the objects should be rearranged in a human-like way). Next, the objects in the initial and generated images are matched using their CLIP visual features. Poses are estimated by aligning their segmentation masks. The robot rearranges the scene based on the estimated poses to create the generated arrangement. 

\begin{table*}[!ht]
\begin{threeparttable}
    \caption{Pretrained Models for Robotics}
    \label{table:robotics_pretrained_models}
    \begin{tabular}{C{0.12\textwidth}C{0.22\textwidth}C{0.07\textwidth}C{0.14\textwidth}C{0.08\textwidth}C{0.22\textwidth}}
    \toprule
    Paper & Backbone & Size (Parameters) & Pretrained Task & Inference Speed & Hardware \tnote{*}\\
    \midrule
    \rowcolor{mygray}
    RoboCat \cite{Bousmalis2023RoboCat} & decoder-only transformer & 1.18B &  manipulation & 10-20Hz& \\ 
    Gato \cite{reed2022generalist}  &decoder-only transformer &1.2B  & generalist agent & 20Hz & 4 days on 16x16 TPU v3 slice\\
    \rowcolor{mygray}
    PaLM-E-562B \cite{driess2023palme}  & decoder-only transformer& 562B & 1Hz for Language subgoals + 5Hz low-level control policies & 5-6Hz& runs on multi-TPU cloud service\\
    ViNT \cite{shah2023vint} & EfficientNet+ decoder transformer  & 31M & visual navigation & 4Hz & variety of GPU configurations is used including 2×4090, 3×Titan Xp, 4×P100, 8×1080Ti, 8×V100, and 8×A100\\
    \rowcolor{mygray}
    VPT \cite{baker2022video} &a temporal convolution layer, a ResNet 62 image
    processing stack, and residual unmasked attention layers, &0.5B  & embodied agent in Minecraft & 20Hz & 9 days on 720 V100 GPUs\\
    RT-1 \cite{RT12023} & Conditioned EfficientNet + TokenLearner + decoder-only transformer &35M & real-world robotics tasks & 3Hz & \\
    \rowcolor{mygray}
    RT-2 \cite{rt22023arxiv} & PaLI-X & 55B & real-world robotics tasks & 1-3Hz &  runs on multi-TPU cloud service\\
    RT-2-X \cite{padalkar2023rtx}& ViT and Language model UL2\cite{UL2}  & 55B & real-world robotics tasks & 1-3Hz & runs on multi-TPU cloud service\\
    \rowcolor{mygray}
    LIV \cite{ma2023liv} & CLIP &  & reward learning & 15Hz & 8 NVIDIA V100 GPUs \\
    SMART \cite{sun2023smart} & decoder-only transformer & 11M & bidirectional dynamics prediction and masked hindsight control &  1 Hz & 8 Nvidia V100 GPUs \\
    \rowcolor{mygray}
    COMPASS \cite{DBLP:conf/iros/MaVWGSMK22} & 3D-Resnet encoder &  20M & Contrastive loss & 30 Hz & 8 Nvidia V100 GPUs \\
    PACT \cite{Bonatti2023} & decoder-only transformer & 12M & forward dynamics and next action prediction &  10 Hz (edge) / 50 Hz & Nvidia Xavier NX (edge) / 8 Nvidia V100 GPUs \\
    \bottomrule
    \end{tabular}    
    \begin{tablenotes}\footnotesize
    \item[*] Empty fields in the table denote no data is reported.
    \end{tablenotes}
\end{threeparttable}
\end{table*}

In Table \ref{table:robotics_pretrained_models} some robotic-specific foundation models are reported along with information about their size and architecture, pretrained task, inference time, and hardware setup. 

\section{Perception}\label{Perception}

Robots interacting with their surrounding environments receive raw sensory information in different modalities such as images, video, audio, and language. This high-dimensional data is crucial for robots to understand, reason, and interact in their environments. Foundation models, including those that have been developed in the vision and NLP domains, are promising tools for converting these high-dimensional inputs into abstract, structured representations that can be more easily interpreted and manipulated. Particularly, multi-modal foundation models enable robots to integrate different sensory inputs into a unified representation encompassing semantic, spatial, temporal, and affordance information. These multi-modal models reflect cross-modal interactions, often by aligning elements across modalities to ensure coherence and correspondence. For example, text and image data are aligned for image captioning tasks. This section will explore a range of tasks related to robot perception that are improved through aligning modalities using foundation models, with a focus on vision and language. There is an extensive body of literature studying multi-modality in the machine learning community, and an interested reader is referred to the survey paper \cite{Morency2019} that presents a taxonomy of multi-modal learning. We focus on applications of multi-modal models to robotics. 

\subsection{Open-Vocabulary Object Detection and 3D Classification}
\subsubsection{Object Detection}
Zero-shot object detection allows robots to identify and locate objects they have never encountered previously. 
Grounded Language-Image Pre-training (GLIP)~\cite{GLIP2022} integrates object detection and grounding by redefining object detection as phrase grounding. This reformulation enables the learning of a visual representation that is both language-aware and semantically rich at the object level. In this framework, the input to the detection model comprises not only an image but also a text prompt that describes all the potential categories for the detection task. To train GLIP, a dataset of 27 million grounding instances was compiled, consisting of 3 million human-annotated pairs and 24 million image-text pairs obtained by web crawling. The results of the study demonstrate the remarkable zero-shot and few-shot transferability of GLIP to a wide range of object-level recognition tasks. Recently, PartSLIP~\cite{liu2023partslip} demonstrated that GLIP can be used for low-shot part segmentation on 3D objects. PartSLIP renders a 3D point cloud of an object from multiple views and combines 2D bounding boxes in these views to detect object parts. To deal with noisy 2D bounding boxes from different views, PartSLIP runs a voting and grouping method on super points from 3D, assigns multi-view 2D labels to super points, and finally groups super points to obtain a precise part segmentation. To enable few-shot learning of 3D part segmentation, prompt tuning, and multi-view feature aggregation are proposed to improve performance.

OWL-ViT~\cite{minderer2022simple} is an open-vocabulary object detector. OWL-ViT uses a vision transformer architecture with contrastive image-text pre-training and detection end-to-end fine-tuning. Unlike GLIP, which frames detection as a phrase grounding problem with a single text query and limits the number of possible object categories, OWL-ViT can handle multiple text-based or image-driven queries. OWL-ViT has been applied to robot learning for example in VoxPoser~\cite{VoxPoser2023} as the open-vocabulary object detector to find ``entities of interest'' (e.g., vase or drawer handles) and ultimately define value maps for optimizing manipulation trajectories.

Grounding DINO~\cite{liu2023grounding} combines DINO~\cite{caron2021emerging} with grounded pre-training, extending the closed-set DINO model to open-set detection by fusing vision and language. Grounding DINO outperforms GLIP in open-set object detection. This superior performance is mainly due to the transformer architecture of Grounding DINO, which facilitates multi-modal feature fusion at multiple stages. 

\subsubsection{3D Classification}
Zero-shot 3D classifiers can enable robots to classify objects in their environments without explicit training data. Foundation models are strong candidates for performing 3D classification.
PointCLIP~\cite{PointCLIP2022} transfers CLIP’s pre-trained knowledge of 2D images to 3D point cloud understanding by aligning point clouds with text. The authors propose to project each point onto a series of pre-defined image planes to generate depth maps. Then, the CLIP visual encoder is used to encode multi-view features of the point cloud and predict labels in natural language for each view. The final prediction for the point cloud is computed via weighted aggregation of the predictions for each view. PointBERT \cite{yu2021pointbert} uses a transformer-based architecture to extract features from point clouds, generalizing the concept of BERT into 3D point clouds.   

Unlike PointCLIP which converts the task of matching point clouds and text to image-text alignment, ULIP~\cite{xue2022ulip,xue2023ulip2} is a Unified representation of Language, Images, and Point clouds for 3D understanding. It achieves this by pre-training with object triplets (image, text, point cloud). The model is trained using a small number of automatically synthesized triplets from ShapeNet55~\cite{shapenet2015}, which is a large-scale 3D model repository. ULIP uses CLIP as the vision-language model. During pretraining, the CLIP model is kept frozen and a 3D encoder is trained by aligning the 3D features of an object with its associated textual and visual features from CLIP using contrastive learning. The pretraining process allows ULIP to learn a joint embedding space where the three modalities are aligned. One of the major advantages of ULIP is that it can substantially improve the recognition ability of 3D backbone models. This is because the pretraining process allows ULIP to learn more robust and discriminative features for each modality, which can then be used to improve the performance of 3D models. Another advantage of ULIP is that it is agnostic to the 3D model architecture, and thus can be easily integrated into the pretraining process of existing 3D pipelines. ULIP adopts masked language modeling from BERT to 3D by tokenizing 3D patches randomly masking out 3D tokens and predicting them back during pretraining. ULIP~\cite{xue2022ulip,xue2023ulip2} has shown that the performance of recognition capability of models such as PointBERT can be improved by using a unified multimodal representation of ULIP.  
 
\subsection{Open-Vocabulary Semantic Segmentation} Semantic segmentation classifies each pixel in an image into semantic classes. This provides fine-grained information about object boundaries and locations within an image and enables embodied agents to understand and interact with the environment at a more granular level. Several works explore how foundation models such as CLIP can enhance the generalizability and flexibility of semantic segmentation tasks.

LSeg is a language-driven semantic segmentation model~\cite{li2022languagedriven} that associates semantically similar labels to similar regions in an embedding space. LSeg uses a text encoder based on the CLIP architecture to compute text embeddings and an image encoder with the underlying architecture of Dense Prediction Transformer (DPT)~\cite{Ranftl_2021_ICCV}. Similar to CLIP, LSeg creates a joint embedding space using text and image embeddings. LSeg freezes the text encoder at training time and trains the image encoder to maximize the correlation between the text embedding and the image pixel embedding of the ground-truth pixel class. It allows users to arbitrarily shrink, expand, or rearrange the label set (with unseen categories) for any image at test time.  

Segment Anything Model (SAM)~\cite{kirillov2023segment} introduces a framework for promptable segmentation consisting of the task definition for promptable segmentation, a segmentation foundation model (the Segment Anything Model, or SAM), and a data engine.  SAM adapts a pretrained Vision Transformer from Masked Auto-Encoder (MAE)~\cite{he2022masked} as an image encoder while using a text encoder from CLIP~\cite{radford2021learning} for sparse prompts (points, boxes, and text) and a separate dense prompt encoder for masks. In contrast to other foundation models that are trained in an unsupervised manner on web-scale data, SAM is trained using supervised learning with data engines that help scale the number of available annotations. Along with the model, the authors released the Segment Anything 1 Billion (SA-1B) dataset. It consists of 11M images and 1.1B segmentation masks. In this work, the authors conducted experiments on five zero-shot transfer tasks, including point-valid mask evaluation, edge detection, object proposal, instance segmentation, and text-to-mask. The system's composable design, facilitated by prompt engineering techniques, enables a broader range of applications compared to systems trained specifically for fixed task sets. However, one limitation of this work that is particularly relevant to robotic applications is that SAM cannot run in real-time.   

FastSAM~\cite{zhao2023fast} and MobileSAM~\cite{mobile_sam} achieve comparable performance to SAM at faster inference speeds. The Track Anything Model (TAM)~\cite{yang2023track} combines SAM and XMem~\cite{cheng2022xmem}, an advanced video object segmentation (VOS) model, to achieve interactive video object tracking and segmentation. Anything-3D~\cite{shen2023anything} employs a collection of visual-language models and SAMs to elevate objects into the realm of 3D. It uses BLIP~\cite{li2022blip} to generate textual descriptions while using SAM to extract objects of interest from visual input. Then, Anything-3D lifts the extracted objects into a Neural Radiance Field (NeRF) \cite{mildenhall2020nerf} representation using a text-to-image diffusion model, enabling their integration into 3D scenes. 

Amidst these remarkable advancements, achieving fine-grained detection with real-time performance still remains challenging. 
For example, LSeg~\cite{li2022languagedriven} reports failure cases related to misclassification, when the test time input labels do not include the true label for the pixel, and the model thus assigns the highest probability to the closest label. Another failure case occurs when multiple labels can be correct for a particular pixel, and the model must classify it as just one of the categories. For example ``window'' and ``house'' may both be defined as labels, but during inference, a pixel representing a ``window'' may be labeled instead as ``house''. SAM also does not provide precise segmentation for fine structures and often fails to produce crisp boundaries. All models that use SAM as a sub-component may encounter similar limitations. 
In the future, fine-grained semantic segmentation models that can assign multiple labels to a pixel when there are multiple correct descriptions should be considered. Additionally, developing models that can run in real-time will be critical for robotics applications.

\subsection{Open-Vocabulary 3D Scene and Object Representations}
Scene representations allow robots to understand their surroundings, facilitate spatial reasoning, and provide contextual awareness. Language-driven scene representations align textual descriptions with visual scenes, enabling robots to associate words with objects, locations, and relationships. In this section, we study recent works that use foundation models to enhance scene representations.  
\subsubsection{Language Grounding in 3D Scene}
Language grounding refers to combining geometric and semantic representations of an environment. One type of representation that can provide an agent with a strong geometric prior is an implicit representation. One example of an implicit representation is a Neural Radiance Field (NeRF)~\cite{mildenhall2020nerf,sun2023nerf,sun2023aria}. NeRF creates high-quality 3D reconstructions of scenes and objects from a set of 2D images captured from different viewpoints (without the need for explicit depth information). The NeRF neural network takes camera poses as input and predicts the 3D geometry of the scene as well as color and intensity. Most NeRF-based models memorize the light field in a single environment and are not pre-trained on a large data set, hence they are not foundation models.  However, foundation models such as CLIP can be combined with NeRFs to extract semantic information from an agent's environment. 

 Kerr et al.~\cite{lerf2023} propose language-embedded radiance fields (LERFs) that ground CLIP embeddings into a dense multi-scale 3D field. This results in a 3D representation of the environment that can be queried to produce semantic relevancy maps. The LERF model takes 3D position $(x,y,z)$, viewing direction ($\phi$, $\theta$), and a scaling factor as input and outputs an RGB value, density ($\sigma$), as well as DINO~\cite{caron2021emerging} and CLIP features. The LERF is optimized in two stages: initially, a multi-scale feature pyramid of CLIP embeddings over training views is computed; then, the pyramid is interpolated using the image scale and pixel location to obtain the CLIP embedding; and finally, the CLIP embeddings are supervised through cosine similarity and the RGB and density are supervised using the standard mean squared-error. 

Models such as LERF inherit the shortcomings of CLIP and NeRF. For example, CLIP exhibits difficulty in capturing spatial relationships between objects. In addition, language queries from CLIP can highlight a significant issue similar to the bag-of-words model, which struggles to distinguish terms with opposite sentiments. Also, NeRF relies on known camera poses associated with pre-captured multi-view images. 

In CLIP-Fields~\cite{shafiullah2022clip}, an implicit scene representation $g(x,y,z): \mathcal{R}^3 \rightarrow \mathcal{R}^d$ is trained by decoding a $d$-dimensional latent vector to different modality-specific outputs. The model distills information from pretrained image models by back-projecting the pixel labels to 3D space and training the output heads to predict semantic labels from an open-vocab object detector called Detic, the CLIP visual representation, and one-hot instance labels using a contrastive loss. The scene representation can then be used as a spatial database for segmentation, instance identification, semantic search over space, and 3D view localization from images. 

Another related work is VLMaps~\cite{huang2023visual}, which projects pixel embeddings from LSeg to grid cells in a top-down grid map. This method does not require training and instead directly backprojects pixel embeddings to grid cells and averages the values in overlapping regions. By combining a VLMap with a code-writing LLM, the authors demonstrate spatial goal navigation using landmarks (e.g., move to the plant) or spatial references with respect to landmarks (between the keyboard and the bowl). Semantic Abstraction (SemAbs)~\cite{ha2022semantic} presents another approach for 3D scene understanding by decoupling visual-semantic reasoning and 3D reasoning. In SemAbs, given an RGB-D image of a scene, a semantic-aware 2D VLM extracts 2D relevancy maps for each queried object while semantic-abstracted 3D modules predict the 3D occupancy of each object using the relevancy maps. Because the 3D modules are trained irrespective of the specific object labels, the system demonstrates strong generalization capabilities, including generalization to new object categories and from simulation to the real world.

Current VLMs can reason about 2D images, however, they are not grounded in the 3D world. The main challenge for building 3D VLM foundation models is the scarcity of 3D data. Particularly, 3D data paired with language description is scarce. One strategy to circumvent this issue is to take advantage of 2D models trained on large-scale data to supervise 3D models. For instance, the authors of FeatureNeRF \cite{ye2023featurenerf} propose to learn 3D semantic representations by distilling 2D vision foundation models (i.e., DINO or Latent Diffusion) into 3D space via neural rendering. FeatureNeRF predicts a continuous 3D semantic feature volume from a single or few images which can be used for downstream tasks such as key-point transfer or object part co-segmentation.   

In 3D-LLM~\cite{3dllm}, the authors propose to use 2D VLMs as backbones to train a 3D-LLM that can take 3D representations (i.e., 3D point clouds with their features) as inputs and accomplish a series of diverse 3D-related tasks. The 3D features are extracted from 2D multi-view images and mapped to the feature space of 2D pretrained VLMs. To overcome 3D data scarcity, the authors propose an efficient prompting procedure for ChatGPT to generate 3D-language data encompassing a diverse set of tasks. These tasks include 3D captioning, dense captioning, 3D question answering, 3D task decomposition, 3D grounding, 3D-assisted dialog, and navigation. Also, to capture 3D spatial information, the authors propose a 3D localization mechanism by 1) augmenting 3D features with position embedding and 2) augmenting LLM vocabularies with 3D location tokens. In the first part, the position embeddings of the three dimensions are generated and concatenated with 3D features. In the second part, the coordinates of the bounding box representing the grounded region are discretized to voxel integers as location tokens $<x_{min}, y_{min}, z_{min}, x_{max}, y_{max}, z_{max}>$. It is important to highlight that, typically, creating 3D representations necessitates the use of 2D multi-view images and camera matrices. These resources are not as readily available as the vast amounts of internet-scale text and image data that current foundation models are trained on.

\subsubsection{Scene Editing}
When an embodied agent relies on an implicit representation of the world, the capability to edit and update this representation enhances the robot's adaptability. For instance, consider a scenario where a robot utilizes a pretrained NeRF model of an environment for navigation and manipulation. If a portion of the environment changes, being able to adjust the NeRF without retraining the model from scratch saves time and resources. 

In the case of NeRFs, Wang et al.~\cite{wang2022clip} propose a text and image-driven method for manipulating NeRFs called CLIP-NeRF. This approach uses CLIP to disentangle the dependence between shape and appearance in conditional neural radiance fields. CLIP-NeRF facilitates the editing of the shape and appearance of NeRFs using either image or text prompts. It is composed of two modules: the disentangled conditional NeRF and CLIP-driven manipulation. The former takes the positional encoding $\gamma(x,y,z)$, a shape code $z_s$, viewing direction $v(\phi,\theta)$, and appearance code $z_a$ as an input and outputs color and density. The disentanglement is achieved using a deformation network that is appended as input to the traditional NeRF MLP that produces density, and by taking the output from this MLP and concatenating it with an appearance code to attain the color value. The CLIP-driven manipulation module takes an image example or text prompt as an input and outputs a shape deformation $\Delta z_s$ and an appearance deformation $\Delta z_a$ from shape mapping and appearance mapping MLPs respectively. These deformation values aim to perturb the shape code and appearance code in the disentangled conditional NeRF module to produce the desired output.

A key limitation of the CLIP-NeRF approach is that prompting can impact the entire scene rather than a selected region. For example, prompting to change the color of a flower's petals might also impact the shape and color of its leaves. To address this limitation, Kobayashi et al. propose to train distilled feature fields (DFFs)~\cite{kobayashi2022decomposing} and then manipulate DFFs through query-based scene decomposition and editing. 
Pre-trained 2D VLMs (such as LSeg~\cite{li2022languagedriven} and DINO~\cite{caron2021emerging}) are employed as teacher networks and distilled into 3D distilled feature fields via volume rendering.  
Editing is achieved by alpha compositing the density and color values of the two NeRF scenes.
When combined with CLIP-NeRF, this method enables CLIP-NeRF to selectively edit specific regions of multi-object scenes.   A similar approach was explored by Tschernezki et al. in  \cite{tschernezki2022neural} where the authors show that enforcing the 3D consistency of features in the NeRF embedding improved segmentation performance compared to using features from the original 2D images.

Another approach to more controlled 3D scene editing is to use structured 3D scene representations. Nerflets~\cite{zhang2023nerflets} represent a 3D scene as a combination of local neural radiance fields where each maintains its own spatial position, orientation, and dimension. Instead of employing a single large MLP to predict colors and densities as standard NeRF, individual Nerflets are combined to predict these values, modulated by their weights. After optimizing posed 2D images and segmentations, Nerflets reflect the decomposed scene and support more controlled editing.

One application of image editing in robotics is for data augmentation during policy learning. ROSIE~\cite{yu2023scaling} use the Imagen editor~\cite{wang2023imagen} to modify training images to add additional distractors and unseen objects and backgrounds to train robust imitation learning policies. GenAug~\cite{chen2023genaug} similarly generates images with in-category and cross-category object substitutions, visual distractors, and diverse backgrounds. The CACTI~\cite{mandi2022cacti} pipeline includes a step in-painting different plausible objects via Stable-Diffusion~\cite{LatentDiffusion2022} onto training images. These approaches generate photorealistic images for training robust policies; however, generating images with sufficient diversity while also maintaining physical realism, e.g. for object contacts, remains a challenge. Existing approaches use learned or provided masks to specify areas of the image to keep, or heuristics based on the particular robotic task.

Another direction is to use generative models to define goal images for planning. DALL-E-Bot~\cite{dall_e_bot2023} uses DALL-E 2 to define a goal image of human-like arrangements from observations. 

\subsubsection{Object Representations}
Learning correspondences between objects can facilitate manipulation by enabling skill transfer from trained objects to novel object instances in known categories or novel object categories at test time. Traditionally, object correspondences have been learned using strong supervision such as keypoints and keyframes. Neural descriptor fields (NDFs)~\cite{simeonov2022neural} remove the need for dense annotation by leveraging layer-wise activations from an occupancy network; however, this approach still requires many training shapes for each target object category. Additional works have started to build object representations directly from image features of pretrained vision models.

Feature Fields for Robotic Manipulation (F3RM)~\cite{shen2023distilled} builds on DFF to develop scene representations that support finding corresponding object regions. F3RM uses a similar feature representation for 6-DoF poses relative to objects (e.g., a grasp on the handle of the mug) to NDF. Besides allowing corresponding 6-DoF poses to be found from a few demonstrations, the pose embeddings can also be directly compared to text embeddings from CLIP to leverage language guidance (e.g., pick up the bowl). Correspondences between objects have also been directly extracted from DINO features \cite{goodwin2023you} without training. This method first extracts dense ViT feature maps of two objects using multiple views. Similar regions on the two objects are found by computing the cyclical distance metric \cite{goodwin2022zero} on the feature maps. With the 2D patch correspondences, a 7-D rigid body transform (i.e., a SO(3) pose, a translation, and a scaling scalar) between the objects can be solved together with RANSAC and Umeyama's method \cite{umeyama1991least}.

\subsection{Learned Affordances}
Affordances refer to the potential of objects, environments, or entities to offer specific functions or interactions to an agent. They can include actions such as pushing, pulling, sitting, or grasping. Detecting affordances bridges the gap between perception and action.

Affordance Diffusion~\cite{ye2023affordance} synthesizes complex interactions of e.g. an articulated hand with a given object. Given an RGB image, Affordance Diffusion aims to generate images of human hands for hand-object interaction (HOI). The authors propose a two-step generative approach based on large-scale pretrained diffusion models based on where to interact (layout) and how to interact (content). The layout network generates a 2D spatial arrangement of hand and object. The content network then synthesizes images of a hand grasping the object conditioned on the given object and the sampled HOI layout. Affordance Diffusion outputs both the hand articulation and approach orientation. 

Vision-Robotic Bridge (VRB)~\cite{bahl2023affordances} trains a visual affordance model on internet videos of human behavior. Particularly, it estimates the likely location and manner in which a human interacts within a scene. This model captures the structural information of these behavioral affordances. The authors seamlessly integrate the affordance model with four different robot learning paradigms. Firstly, they apply offline imitation learning, where the robot learns by imitating the observed human interactions from the videos. Secondly, they use exploration techniques to enable the robot to actively discover and learn new affordances in its environment.
Thirdly, the authors incorporate goal-conditioned learning, allowing the robot to learn how to achieve specific objectives by leveraging the estimated affordances. Finally, they integrate action parameterization for reinforcement learning, enabling the robot to learn complex behaviors by optimizing its actions based on the estimated affordances.

\subsection{Predictive Models}
Predictive dynamics models, or world models, predict how the state of the world changes given particular agent actions, that is, they attempt to model the state transition function of the world~\cite{sun2021adversarial}. When applied to visual observations, dynamics modeling can be formulated as a video prediction problem~\cite{sun2023connected,sun2022self}. While video generation and prediction, particularly over long horizons, is a longstanding challenge with many prior efforts, recent models based on vision transformers and diffusion models have demonstrated improvements~\cite{janner2022diffuser,sun2023conformal}. For instance, the Phenaki model~\cite{Villegas2022PhenakiVL} generates variable length video up to minutes in length conditioned on text prompts. 

Several approaches apply these models to robotics in the literature. Note that while learned dynamics or world models in robotics have been explored in constrained or smaller-data regimes, we focus in this section on works that train on a diversity or volume of data that is characteristic of foundation models. One strategy is to learn an action-conditioned model that may be used directly for downstream planning by optimizing an action sequence~\cite{dasari2019robonet}, i.e. performing model-predictive control, or for policy learning via training on simulated rollouts. 
One example is the GAIA-1 model which generates predictions of driving video conditioned on arbitrary combinations of video, action, and text~\cite{hu2023gaia1}. It was trained on $4700$ hours of proprietary driving data. Another approach is to use a video prediction model to generate a plan of future states, and then learn a separate goal-conditioned policy or inverse dynamics model to infer control actions based on the current and target state. One line of work instantiates this by combining text-conditioned video diffusion models with image-goal-conditioned policies to solve manipulation tasks in simulated and real tabletop settings~\cite{du2023learning}. This approach has been extended to longer-horizon object manipulation tasks by using the PaLM-E VLM to break down a high-level language goal into smaller substeps, leveraging feedback between the VLM and video generation models~\cite{du2023vlp}. 

Another example is COMPASS \cite{DBLP:conf/iros/MaVWGSMK22}, which first constructs a comprehensive multimodal graph to capture crucial relational information across diverse modalities. The graph is then used to construct a rich spatio-temporal and semantic representation. Pretrained on the TartanAir multimodal dataset, COMPASS was demonstrated to address multiple robotic tasks including drone navigation, vehicle racing, and visual odometry.

\section{Embodied AI}\label{EmbodiedAI}

Recently, researchers have shown that the
the success of LLMs can be extended to embodied AI domains~\cite{Ahn2022DoAI,zeng2022socraticmodels,huang2022inner,codeaspolicies2022}, where ``embodied'' typically refers to a virtual embodiment in a world simulator, not a physical robot embodiment. Statler~\cite{yoneda2023statler} is a framework that endows LLMs with an explicit representation of the world state as a form of “memory” that is maintained over time. Statler uses two instances of general LLMs: a world-model reader and a world-model writer, that interface with and maintain the world state. Statler improves the ability of existing LLMs to reason over longer time horizons without the constraint of context length.

Large Scale Language Models (LSLMs) have exhibited strong reasoning ability and the ability to adapt to new tasks through in-context learning. Dasgupta et al.~\cite{dasgupta2022collaborating} combine these complementary abilities in a single system consisting of three parts: a Planner, an Actor, and a Reporter. The Planner is a pretrained language model that can issue commands to a simple embodied agent (the Actor), while the Reporter communicates with the Planner to inform its next command. Mu et al.~\cite{mu2023embodiedgpt} build EgoCOT, a dataset consisting of carefully selected videos from the Ego4D dataset, along with corresponding high-quality language instructions. EmbodiedGPT~\cite{mu2023embodiedgpt} utilizes prefix adapters to augment the 7B language model's capacity to generate high-quality planning, training it on the EgoCOT dataset to avoid overly divergent language model responses. Comprehensive experiments were conducted, demonstrating that the model effectively enhances the performance of embodied tasks such as Embodied Planning, Embodied Control, Visual Captioning, and Visual Q\&A. Embodied agents should autonomously and endlessly explore the environment. They should actively seek new experiences, acquire new skills, and improve themselves. 

The game of Minecraft~\cite{engelbrecht2014transforming} provides a platform for designing intelligent agents capable of operating in the open world. MineDojo~\cite{fan2022minedojo} is a framework for developing generalist agents in the game of Minecraft. MineDojo offers thousands of open-ended and language-prompted tasks, where the agent can navigate in a progressively generated 3D environment to mine, craft tools, and build structures. As part of this work, the authors introduce MiniCLIP, a video-language model that learns to capture the correlations between a video clip and its time-aligned text that describes the video. The MineCLIP model, trained on YouTube videos, can be used as a reward function to train the agent with reinforcement learning. By maximizing this reward function, it incentivizes the agent to make progress toward solving tasks specified in natural language.

Voyager~\cite{wang2023voyager} introduces an LLM-powered embodied lifelong learning agent in the realm of Minecraft. Voyager uses GPT-4 to continuously explore the environment. It interacts with GPT-4 through in-context prompting and does not require model parameter fine-tuning. Exploration is maximized by querying GPT-4 to provide a stream of new tasks and challenges based on the agent's history interactions and current situations. Also, the iterative prompting mechanism generates code as the action space to control the Minecraft agent. Iterative prompting incorporates environment feedback provided by Minecraft, execution errors, and a self-verification scheme. For self-verification, GPT-4 acts as a critic by checking task success and providing suggestions for task completion in the case of failure. The GPT-4 critic can be replaced by a human critic to provide on-the-fly human feedback during task execution. Ghost in the Minecraft (GITM)~\cite{zhu2023ghost} leverages LLM to break down goals into sub-goals and map them to structured actions for generating control signals. GITM consists of three components: an LLM Decomposer, an LLM Planner, and an LLM Interface. The LLM Decomposer is responsible for dividing the given Minecraft goal into a sub-goal tree. The LLM Planner then plans an action sequence for each sub-goal. Finally, the LLM Interface executes each action in the environment using keyboard and mouse operations. 

Reinforcement learning in embodied AI virtual environments has the potential to improve the capabilities of real-world robotics by providing efficient training and optimizing control policies in a safe and controlled setting. Reward design is a crucial aspect of RL that influences the robot's learning process. Rewards should be aligned with the task's objective and guide the robot to achieve the desired task. Foundation models can be leveraged to design rewards. Kwon et al.~\cite{kwon2023reward} investigate the simplification of reward design by utilizing a large language model (LLM), such as GPT-3, as a proxy reward function. In this approach, users provide a textual prompt that contains a few examples (few-shots) or a description (zero-shot) of the desired behavior.
The proposed method incorporates this proxy reward function within a reinforcement learning framework. Users specify a prompt at the start of the training process. During training, the RL agent's behavior is evaluated by the LLM against the desired behavior outlined in the prompt, resulting in a corresponding reward signal generated by the LLM. Subsequently, the RL agent employs this reward to update its behavior through the learning process. 

In~\cite{GudingRL2023}, the authors propose a method called Exploring with LLMs (ELLM) that rewards an agent for achieving goals suggested by a language model. The language model is prompted with a description of the agent's current state. Therefore, without having a human in the loop, ELMM guides agents toward meaningful behavior.

Zhang et al.~\cite{zhang2022offline} explore the potential relationship between offline reinforcement learning and language modeling. They hypothesize that RL and LM share similarities in predicting future states based on current and past states, considering both local and long-range dependencies across states. To validate this assumption, the authors pre-train Transformer models on different offline RL tasks and assess their performance on various language-related tasks.
Tarasov et al.~\cite{tarasov2022prompts} present an approach to harness pretrained language models in deep offline reinforcement learning scenarios that are not inherently compatible with textual representations. The authors suggest a method that involves transforming the RL states into human-readable text and performing fine-tuning of the pretrained language model during training with deep offline RL algorithms.

Advances in model architecture (e.g. transformer) for foundation models allow the model to effectively model and predict sequences. To harness the power of these models, some recent studies investigate exploiting these architectures for sequence modeling in RL problems. Reid et al.~\cite{reid2023can} explore the potential of leveraging the sequence modeling formulation of reinforcement learning and examine the transferability of pretrained sequence models across different domains, such as vision and language. They specifically focus on the effectiveness of fine-tuning these pretrained models on offline RL tasks, including control and games. In addition to investigating the transferability of pretrained sequence models, the authors propose techniques to enhance the transfer of knowledge between these domains. These techniques aim to improve the adaptability and performance of the pretrained models when applied to new tasks or domains.

High-level task planning using LLMs is demonstrated in embodied AI environments. Huang et al.~\cite{huang2022language} propose employing pretrained Language Models (LMs) as zero-shot planners. The approach is evaluated in the  VirtualHome~\cite{puig2018virtualhome} environment. In this work, first, an autoregressive LLM such as GPT-3~\cite{brown2020language} or Codex~\cite{chen2021evaluating} is quarried to generate action plans for high-level tasks. Some of these action plans might not be executable by the agent due to ambiguity in language or referring to objects that are not present or grounded in the environment. So, to select the admissible action plans, admissible environment actions, and generated actions by the causal LLM are embedded using a BERT-style LM. Then for each admissible environment action, its semantic distance to the generated action is computed using cousin similarity.  

Chain of thought reasoning and action generation are proposed for embodied agents as well. 
ReAct \cite{yao2023react}
combines reasoning (e.g. chain of thought) and acting (e.g. sequence of action generation) within LLM. 
Reasoning traces enhance the model's ability to deduce, monitor, and revise action plans, along with managing exceptions effectively. Actions facilitate interaction with external resources, like knowledge bases or environments, enabling it to acquire supplementary information. 
ReAct showcases its proficiency across a wide array of language and decision-making tasks, including question-answering and fact verification. It enhances interpretability and trust for users by transparently illustrating the process through which it searches for evidence and formulates conclusions. Unlike prior methods that depend on a singular chain-of-thought, ReAct engages with a Wikipedia API for pertinent information retrieval and belief updating. This strategy effectively mitigates the issues commonly associated with chain-of-thought reasoning, such as hallucination and error propagation.

VPT~\cite{baker2022video} presents video pretraining in which the agent learns to act by watching unlabeled online videos. It is shown that an inverse dynamic model can be trained with a small labeled dataset and the model can be used to label a huge unlabeled data of the internet. Videos of people who have played Minecraft are used to train an embodied AI agent to play Minecraft. The model exhibits zero-shot performance and can be fine-tuned for more complex skills using imitation learning or reinforcement learning. The VPT model is trained with a standard behavioral cloning loss (\ref{eq:BC}) (negative log-likelihood) while the actions are drawn from the inverse dynamic model. 

\subsection{Generalist AI}
A long-standing challenge in robotics research is deploying robots or embodied AI agents in a variety of non-factory real-world applications, performing a range of tasks.
To make generalist robots that can operate in diverse environments with diverse tasks, some researchers have proposed generative simulators for robot learning. For example, Generative Agents~\cite{generativeAI2023} discusses how generative agents can produce realistic imitations of human behavior for interactive applications, creating a miniature community of agents similar to those found in games like The Sims. The authors connect their architecture with the ChatGPT large language model to create a game environment with 25 agents. The study includes two evaluations, a controlled evaluation and an end-to-end evaluation, which demonstrate the causal effects of the various components of their architecture. Xian et al.~\cite{generalistRobot2023}, authors propose a fully automated generative pipeline, known as a generative simulation for robot learning, which utilizes models to generate diverse tasks, scenes, and training guidance on a large scale. This approach can facilitate the scaling up of low-level skill learning, ultimately leading to a foundational model for robotics that empowers generalist robots.

An alternative method for developing generalist AI involves using generalizable multi-modal representations.
Gato~\cite{reed2022generalist} is a generalist agent that works as a multi-modal, multi-task, multi-embodiment generalist policy. Using the same neural network with the same set of weights, Gato can sense and act with different embodiments in various environments across different tasks. Gato can play Atari, chat, caption images, stack blocks with a real robot arm, navigate in a 3D simulated environment, and more. Gato is trained on 604 different tasks with various modalities, observations, and actions. In this setting, language acts as a common grounding across different embodiments. Gato has 1.2B parameters and is trained offline in a supervised way. Positioned at the confluence of representation learning and reinforcement learning (RL), RRL~\cite{shah2021rrl} learns behaviors directly from proprioceptive inputs. By harnessing pre-trained visual representations, RRL is able to learn from visual inputs, which typically pose challenges in conventional RL settings.

\subsection{Simulators}

High-quality simulators or benchmarks are crucial for robotics development. Hence, we put the ``simulator'' section here to highlight its essential role. To facilitate generalization from simulation to the real world, Gibson~\cite{xiazamirhe2018gibsonenv} emphasizes real-world perception for embodied agents. To bridge the gap between simulation and real-world, iGibson~\cite{li2021igibson} and BEHAVIOR-1K~\cite{li2022behaviork} further support the simulation of a more diverse set of household tasks and reach high levels of simulation realism. As a simulation platform for research in Embodied AI,  Habitat~\cite{habitat19iccv} consists of Habitat-Sim and Habitat-API. Habitat-Sim can achieve several thousand frames per second (fps) running single-threaded. 
Rather than modeling into low-level physics, Habitat-Lab~\cite{szot2021habitat}, is a high-level library for embodied AI, giving a modular framework for end-to-end development. It facilitates the definition of embodied AI tasks, such as navigation, interaction, instruction following, and question answering. Additionally, it enables the configuration of embodied agents, encompassing their physical form, sensors, and capabilities. The library supports various training methodologies for these agents, including imitation learning, reinforcement learning, and traditional non-learning approaches like the SensePlanAct pipelines. Furthermore, it provides standard metrics for evaluating agent performance across these tasks. In line with this, the recent release of Habitat 3.0~\cite{puig2023habitat3} further expands these capabilities.

Similarly, RoboTHOR~\cite{deitke2020robothor} serves as a platform for the development and evaluation of embodied AI agents, offering environments in both simulated and physical settings. Currently, RoboTHOR includes a training and validation set comprising 75 simulated scenes. Additionally, there are 14 scenes each for test-dev and test-standard in the simulation, with corresponding physical counterparts. Key features of RoboTHOR include its reconfigurability and benchmarking capabilities. The physical environments are constructed using modular, movable components, enabling the creation of diverse scene layouts and furniture configurations in a single physical area.
Another simulator, VirtualHome~\cite{puig2018virtualhome}, models complex activities that occur in a typical household. It supports program descriptions for a variety of activities that happen in people’s homes. Huang et al.~\cite{huang2022inner} use VirtualHome to evaluate the robot planning ability with language models.
These simulators have the potential to be applied for evaluating LLMs on robotics tasks.

\section{Challenges and Future Directions}~\label{challenges and Future Directions}
In this section, we examine challenges related to integrating foundation models into robotics settings. We also explore potential future avenues to address some of these challenges.
\subsection{Overcoming Data Scarcity in Training Foundation Models for Robotics}
One main challenge is that compared to the internet-scale text and image data that large models are trained on, robotic-specific data is scarce. We discuss various techniques to overcome data scarcity. For example, to scale up robot learning, some recent works suggest the use of play data instead of expert data for imitation learning. Another technique is data augmentation using in-painting techniques.    
\subsubsection{Scaling Robot Learning Using Unstructured Play Data and Unlabeled Videos of Humans}
Language-conditioned learning such as language-conditioned behavioral cloning, or language-conditioned affordance learning requires having access to large annotated datasets. To scale up learning, in Play-LMP~\cite{lynch2019play}, the authors suggest using teleoperated human-provided play data instead of fully annotated expert demonstrations. Play data is unstructured, unlabeled, cheap to collect, but rich. Collecting play data does not require scene staging, task segmenting, or resetting to an initial state. Also, in MimicPlay~\cite{wang2023mimicplay} a goal-conditioned trajectory generation model is trained based on human-play data. The play data includes unlabeled video sequences of humans interacting with the environment with their hands. Recently works such as~\cite{mees23hulc2} have shown a very small percentage (as little as 1\%) of language-annotated data is needed to train a visuo-lingual affordance model for robot manipulation tasks. 
\subsubsection{Data Augmentation using Inpainting}
Collecting robotics data requires the robot to interact with the real physical world. This data collection process can be associated with significant costs and potential safety concerns. One way to tackle this challenge is to use generative AI such as text-to-image diffusion models for data augmentation. 
For example, ROSIE (Scaling Robot Learning with Semantically Imagined Experience)~\cite{yu2023scaling} presents a diffusion-based data augmentation. Given a robot manipulation dataset, they use inpainting to create 
various unseen objects, backgrounds, and distractors with textual guidance. One important challenge for these methods is developing inpainting strategies that can generate sufficient semantically and visually diverse data, while at the same time ensuring that this data is physically feasible and accurate. For instance, using inpainting to modify an image of an object within a robot's gripper may result in an image with a physically unrealistic grasp, leading to poor downstream training performance. Additional investigation into generative foundation models that are evaluated not only for visual quality but also for physical realism may improve the generality of these methods.

\subsubsection{Overcoming 3D Data Scarcity for Training 3D Foundation Models}
Currently, multi-modal Vision-and-Language Models (VLMs) can analyze 2D images, but they lack a connection to the 3D world, which encompasses 3D spatial relationships, 3D planning, 3D affordances, and more. The primary obstacle in developing foundational 3D VLM models lies in the scarcity of 3D data, especially data that is paired with language descriptions. As discussed, language-driven perception tasks such as language-driven 3D scene representation, language-driven 3D scene editing, language-driven 3D scene or shape generation, language-driven 3D classification, and affordance prediction require access to 3D data or multi-view images with camera matrices which are not readily available data types. New datasets or data generation methods need to be created in the future to overcome data scarcity in the 3D domain.

\subsubsection{Synthetic Data Generation via High-Fidelity Simulation}
High-fidelity simulation via gaming engines can provide an efficient means to collect data, especially to solve multimodal and 3D perception tasks on robots. For example, TartanAir \cite{tartanair2020iros}, a dataset for robot navigation tasks,  was collected in \cite{airsim2017fsr} with the presence of moving objects, changing light, and various weather conditions. By collecting data in simulations, it was possible to obtain multi-modal sensor data and precise ground truth labels such as the stereo RGB image, depth image, segmentation, optical flow, camera poses, and LiDAR point cloud. A large number of environments were set up with various styles and scenes, covering challenging viewpoints and diverse motion patterns that are difficult to achieve by using physical data collection platforms. An extension TartanAir-V2 (https://tartanair.org) furthers the dataset by incorporating additional environments and modalities, such as fisheye, panoramas, and pinholes, with arbitrary camera intrinsic and rotations.

\subsubsection{Data Augmentation using VLMs}
Data augmentation can be provided using Visual-Language Models (VLMs). 
In DIAL~\cite{xiao2022robotic}, Data-driven Instruction Augmentation for Language-conditioned control is introduced. DIAL uses VLM to label offline datasets for language-conditioned policy learning. DIAL performs instruction augmentation using VLMs to weakly relabel offline control datasets. DIAL consists of three steps 1) Contrastive fine-tuning of a VLM such as CLIP~\cite{CLIP2021} on a small robot manipulation dataset of trajectories with crowd-sourced annotation, 2) producing new instruction labels by using the fine-tuned VLM to score relevancy of crowd-sourced annotations against a larger dataset of trajectories, 3) training a language-conditioned policy using behavior cloning on both, the original and re-annotated dataset.

\subsubsection{Robot Physical Skills are Limited to Distribution of Skills} 
One key limitation of the existing robot transformers and other related works in robotics is that robot physical skills are limited to the distribution of skills observed within the robot data. Using these transformers, the robot lacks the capability to generate new movements. To address this constraint, an approach involves using motion data from videos that humans performing various tasks. The inherent motion information within these videos can then be employed to facilitate the acquisition of physical skills in robotics.

\subsection{Real Time Performance (High Inference Time of Foundation Models)} 
Another bottleneck for deploying foundation models on robots is the inference time of these models. In Table~\ref{table:robotics_pretrained_models}, the inference time for some of these models is reported. As seen, the inference time for some of the models still needs to be improved for reliable real-time deployment of the robotic systems. As real-time capability is an essential requirement for any robotic system, more research needs to be performed to improve the computational efficiency of foundation models. 

Furthermore, foundation models are most often stored and run in remote data centers, and accessed through APIs that require network connectivity. Many foundation models (e.g., the GPT models, the Dall-E models) can only be accessed this way, while others are usually accessed this way, but can also be downloaded and run locally with sufficient local computing power (such as SAM~\cite{kirillov2023segment}, LLaMA~\cite{Llama}, and DINOv2~\cite{oquab2023dinov2}).  Given this cloud-service paradigm, the latencies and service times in response to an API call for a foundation model depend on the underlying network over which the data is routed and the data center where the computation takes place---factors that are beyond the control of a robot. So network reliability should be taken into account before integrating a foundation model into a robot's autonomy stack. 

For some robotics domains reliance on the network and 3rd party computing may not be a safe or realistic operating paradigm.  In autonomous driving, autonomous aircraft, search and rescue or emergency response applications, and defense applications the robot cannot rely on network connectivity for time-critical perception or control computations. One option is to have a safe fall-back mode that relies on classical autonomy tools using only local computation, that can take over if access to the cloud is interrupted for some reason.  Another potential longer-term solution for network-free autonomy is the distillation of large foundation models into smaller-sized specialized models that run on onboard robot hardware.  Some recent work has attempted this approach (though without an explicit link to robotics) \cite{lin2023awq}. Such distilled models would likely give up some aspect of the full model, e.g. restricting operation to a certain limited context, in exchange for smaller size and faster compute.  This could be an interesting future direction for bringing the power of foundation models to safety-critical robotics systems.     
 
\subsection{Limitations in Multimodal Representation}
Multimodal interaction implicitly assumes that the modality is tokenizable and can be standardized into input sequences without losing information. The Multimodal models provide information sharing between multiple modalities and are some variation of multimodal transformers with cross-modal attention between every pair of inputs. 
In multimodal representation learning, it is assumed that cross-modal interactions and the dimension of heterogeneity between different modalities can all be captured by simple embeddings. In other words, a simple embedding is assumed to be sufficient to identify the modality or for example, how different language is from vision. In the realm of multimodal representation learning, the question of whether a single multimodal model can accommodate all modalities remains an open challenge. 

Additionally, when paired data between a modality and text is available one can embed that modality into text directly. In robotics applications there are some modalities for which sufficient data is not available and to be able to align them with other modalities, they need to be first converted to other modalities and then be used. For example, 3D point cloud data has various applications in robotics but training a foundation model using this type of data is challenging since data is scarce and is not aligned with text. So, one way to overcome this challenge is first converting this 3D point cloud data to other modalities such as images and subsequently images to text as the secondary step of alignment. Then they can be used in foundation model training. As another example, in Socratic models~\cite{zeng2022socraticmodels}, each modality, whether visual or auditory, is initially translated into language, after which language models attempt to respond to these modalities.  

\subsection{Uncertainty Quantification}

How can we provide assurances on the reliability of foundation models when they are deployed in potentially safety-critical robotics applications~\cite{sun2023conformal}? Current foundation models such as LLMs often \emph{hallucinate}, i.e., produce outputs that are factually incorrect, logically inconsistent, or physically infeasible. While such failures may be acceptable in applications where the outputs from the model can be checked by a human in real-time (e.g., as is often the case for LLM-based conversational agents), they are not acceptable when deploying autonomous robots that use the outputs of foundation models in order to act in human-centered environments. Rigorous uncertainty quantification is a key step toward addressing this challenge and safely integrating foundation models into robotic systems. Below, we highlight challenges and recent progress in uncertainty quantification for foundation models in robotics. 

\subsubsection{Instance-Level Uncertainty Quantification} How can we quantify the uncertainty in the output of a foundation model for a \emph{particular} input? As an example, consider the problem of image classification; given a particular image, one may quantify uncertainty in the output by producing a \emph{set} of object labels that the model is uncertain among or a \emph{distribution} over object labels. Instance-level uncertainty quantification can inform the robot's decisions at runtime. For example, if an image classification model running on an autonomous vehicle produces a prediction set $\{ \texttt{Pedestrian, Bicyclist}\}$ representing that it is uncertain whether a particular agent is a pedestrian or a bicyclist, the autonomous vehicle can take actions that consider both possibilities. 

\subsubsection{Distribution-Level Uncertainty Quantification} How can we quantify the uncertainty in the correctness of a foundation model that will be deployed on a \emph{distribution} of possible future inputs? For the problem of image classification, one may want to compute or bound the probability of errors over the distribution of inputs that a robot may encounter when deployed. Distribution-level uncertainty quantification allows us to decide whether a given model is sufficiently reliable to deploy in our target distribution of scenarios. For example, we may want to collect additional data or fine-tune the model if the computed probability of error is too high. 

\subsubsection{Calibration} In order to be useful, estimates of uncertainty (both at the instance-level and distribution level) should be \emph{calibrated}. If we perform instance-level uncertainty quantification using prediction sets, calibration asks for the prediction set to contain the true label with a user-specified probability (e.g., $95\%$) over future inputs. If instance-level uncertainty is quantified using a distribution over outputs, it should be the case that outputs that are assigned confidence $p$ are in fact correct with probability $p$ over future inputs. Similarly, distribution-level uncertainty estimates should bound the true probability of errors when encountering inputs from the target distribution. 

We highlight a subtle, but important, point that is often overlooked when performing uncertainty quantification in robotics: it can be crucial to pay attention to the distinction between Frequentist and Bayesian interpretations of probabilities. In many robotics contexts --- particularly safety-critical ones --- the desired interpretation is often Frequentist in nature. For example, if we produce a bound $\epsilon$ for the probability of collision of an autonomous vehicle, this should bound the actual observed rate of collisions when the vehicle is deployed. Bayesian techniques (e.g., Gaussian processes or Bayesian ensembles) do not necessarily produce estimates of uncertainty that are calibrated in this Frequentist sense (since the estimates depend on the specific prior that is used to produce the estimates). Trusting the resulting uncertainty estimates may lead one astray if the goal is to provide statistical guarantees on the safety or performance of the robotic system when it is deployed. 

\subsubsection{Distribution Shift} An important challenge in performing calibrated uncertainty quantification is \emph{distribution shift}. A foundation model trained on a particular distribution of inputs may not produce calibrated estimates of uncertainty when deployed on a different distribution for a downstream task. A more subtle cause of distribution shift in robotics arises from \emph{closed-loop} deployment of a model. For example, imagine an autonomous vehicle that chooses actions using the output of a perception system that relies on a pretrained foundation model; since the robot's actions influence future states and observations, the distribution of inputs the perception system receives can be potentially very different from the one it was trained on. 

\subsubsection{Case Study: Uncertainty Quantification for Language-Instructed Robots} Recently, there has been exciting progress in performing rigorous uncertainty quantification for language-instructed robots~\cite{knowno2023}. This work proposes an approach called \textsc{KnowNo} for endowing language-instructed robots with the ability to \emph{know when they don't know} and to ask for help or clarification from humans in order to resolve uncertainty. \textsc{KnowNo} performs both instance-level and distribution-level uncertainty quantification in a calibrated manner using the theory of \emph{conformal prediction}. In particular, given a language instruction (and a description of the robot's environment generated using its sensors), conformal prediction is used to generate a prediction set of candidate actions. If this set is a singleton, the robot executes the corresponding action; otherwise, the robot seeks help from a human by asking them to choose an action from the generated set. Using conformal prediction, \textsc{KnowNo} ensures that asking for help in this manner results in a statistically guaranteed level of task success (i.e., distribution-level uncertainty quantification). \textsc{KnowNo} tackles potential challenges with distribution shift by collecting a small amount of calibration data from the target distribution of environments, tasks, and language instructions, and using this as part of the conformal prediction calibration procedure. While \textsc{KnowNo} serves as an example of calibrated instance-level and distribution-level uncertainty quantification for LLMs, future research should also explore assessing and ensuring the reliability of various other foundation models, such as vision-language models, vision-navigation models, and vision-language-action models, commonly employed in robotics. In addition, exploring how Bayesian uncertainty quantification techniques (e.g., ensembling \cite{Young2023, sun2022quantifying}) can be combined with approaches such as conformal prediction to produce calibrated estimates of instance-level and distribution-level uncertainty is a promising direction. 

\subsection{Safety Evaluation}

The problem of \emph{safety evaluation} is closely related to uncertainty quantification. How can we rigorously \emph{test} for the safety of a foundation model-based robotic system (i) before deployment, (ii) as the model is updated during its lifecycle, and (iii) as the robot operates in its target environments? We highlight challenges and research opportunities related to these problems below. 

\subsubsection{Pre-deployment safety tests} Rigorous pre-deployment testing is crucial for ensuring the safety of any robotic system. However, this can be particularly challenging for robots that incorporate foundation models. First, foundation models are trained on vast amounts of data; thus, a rigorous testing procedure should ensure that test scenarios were not seen by the model during training. Second, foundation models often commit errors in ways that are hard to predict \emph{a priori}; thus, tests need to cover a diverse enough range of scenarios to uncover flaws. Third, foundation models such as LLMs are often used to produce open-ended outputs (e.g., a plan for a robot described in natural language). The correctness of such outputs can be challenging to evaluate in an automated manner if these outputs are evaluated in isolation from the entire system. 

The deployment cycle of current foundation models (in non-robotics applications) involves thorough \emph{red-teaming} by human evaluators \cite{openai2023gpt4, ganguli2022red}. Recent work has also considered partially automating this process by using foundation models themselves to perform red-teaming \cite{perez2022red, tong2023mass}. Developing ways to perform red-teaming (both by humans and in a partially automated way) for foundation models in robotics is an exciting direction for future research.

In addition to evaluating the foundation model in isolation, it is also critical to assess the safety of the end-to-end robotic system. Simulation can play a critical role here, and already does so for current field-deployed systems such as autonomous vehicles \cite{kusano2022collision, webb2020waymo}. The primary challenges are to ensure that (i) the simulator has high enough fidelity for results to meaningfully transfer to the real world, and (ii) test scenarios (manually specified, replicated from real-world scenarios, or automatically generated via adversarial methods \cite{ding2023survey}) are representative of real-world scenarios and are diverse enough to expose flaws in the underlying foundation models. In addition, finding ways to augment large-scale simulation-based testing with smaller-scale real-world testing is an important direction for future work. We emphasize the need for performing such testing throughout the lifecycle of a field-deployed robotic system, especially as updates are made to different components (which may interact in unpredictable ways with foundation models).

\subsubsection{Runtime monitoring and out-of-distribution detection} In addition to performing rigorous testing \emph{offline}, robots with foundation model-based components should also perform \emph{runtime monitoring}. This can take the form of \emph{failure prediction} in a given scenario, which can allow the robot to deploy a safety-preserving fallback policy \cite{luo2022sample,farid2022failure, farid2023task, hsu2023sim, hsu2023safety}. Alternately, the robot can perform \emph{out-of-distribution (OOD) detection} using experiences collected from a small batch of scenarios in a novel distribution \cite{farid2022task, greenberg2021detecting, cai2020real, sinha2022system}; this can potentially trigger the robot to cease its operations and collect additional training data in the novel distribution in order to re-train its policy. Developing techniques that perform runtime monitoring and OOD detection with statistical guarantees on false positive/negative error rates in a data-efficient manner remains an important research direction. 

\subsection{Using Existing Foundation Models as Plug-and-Play or Building New Foundation Models for Robotics}
To incorporate foundation models into robotics, either the existing pretrained large models can be employed as plug-and-play or new foundation models can be built using robotics data. Using foundation models as plug-and-play refers to integrating foundation models into various applications without the need for extensive customization. A large body of the existing literature on foundation models in robotics is centered around the use of foundation models from other domains such as language or vision as plug-and-play. The plug-and-play approach simplifies and facilitates the integration of recent AI advances into the robotics domain. While employing these models as plug-and-play offers a convenient way to harness the power of AI and provide rapid implementation, versatility, and scalability, they are not always customized to specific applications. When specific domain expertise is needed, it is necessary to build a foundation model from scratch or fine-tune the existing models. Building a foundation model from scratch is resource-intensive and demands significant computational power. However, it provides fine-grained control over the architecture, training parameters, and overall behavior.

\subsection{High Variability in Robotic Settings}
Another challenge is the high variability in robotic settings. Robot platforms are inherently diverse with different physical characteristics, configurations, and capabilities. Real-world environments that robots operate in are also diverse and uncertain with a wide range of variations. Due to all these variabilities, robotic solutions are usually tailored to specific robot platforms with specific layouts, environments, and objects for specific tasks. These solutions are not generalizable across various embodiments, environments, or tasks. Hence, to build general-purpose pretrained robotic foundation models, a key factor is to pre-train large models that are task-agnostic, cross-embodiment, and open-ended and capture diverse robotic data. In ROSIE~\cite{yu2023scaling} a diverse dataset is generated for robot learning by performing inpainting of various unseen objects, backgrounds, and distractors with semantic textual guidance. To overcome variability in robotic settings and improve generalization, another solution as ViNT~\cite{shah2023vint} presents is to train foundation models on diverse robotic data across various embodiments. RT-X~\cite{padalkar2023rtx} also investigates the possibility of training large cross-embodied robotic models in the domain of robotic manipulation. RT-X is trained using a multi-embodiment dataset which is created by collecting data from different robot platforms collected through a collaboration between 21 institutions, demonstrating 160266 tasks. RT-X demonstrates transfer across embodiment improves robot capabilities by employing experience from diverse robotic platforms.  

\subsection{Benchmarking and Reproducibility in Robotics Settings}
Another significant obstacle in incorporating foundation models into robotics research is the necessary reliance on real-world hardware experiments. This creates challenges for reproducibility, as replicating results obtained from hardware experiments may necessitate access to the exact equipment employed. Conversely, many recent works have relied on non-physics-based simulators (e.g., ignoring or greatly simplifying contact physics in gasping) that instead focus on high-level, long-term tasks and visual environment models.  Examples of this class of simulators are common and include many of the simulators described above in Sec.~\ref{EmbodiedAI}.  For example the Gibson family of simulators \cite{xiazamirhe2018gibsonenv,li2021igibson}, the Habitat family \cite{szot2021habitat,habitat19iccv, puig2023habitat3}, RobotTHOR \cite{deitke2020robothor}, and VirtualHome \cite{puig2018virtualhome} all neglect low-level physics in favor of simulating higher level tasks with high visual fidelity. 
 This leads to a large sim-to-real gap and introduces variability in real-world performance based on how low-level planning and control modules handle the true physics of the scenario. Even when physics-based simulators are used (e.g., PyBullet or MuJoCo), the absence of standardized simulation settings, computing environments, and a persistent sim-to-real gap impede efforts to benchmark and compare performance across various research endeavors. A combination of open hardware, benchmarking in physics-based simulators, and promoting transparency in experimental and simulation setups can significantly alleviate the challenges associated with reproducibility in the integration of foundation models into robotics research. These practices contribute to the development of a more robust and collaborative research ecosystem within the field.

\section{Conclusion}\label{Conclusion}

Through examination of the recent literature, we have surveyed the diverse and promising applications of foundation models in robotics. We have delved into how these models have enhanced the capabilities of robots in areas such as decision-making, planning and control, and perception. We also discussed the literature on embodied AI and generalist AI, with an eye toward opportunities for roboticists to extend the concepts in that research field to real-world robotic applications. Generalization, zero-shot capabilities, multimodal capabilities, and scalability of foundation models have the potential to transform robotics. However, as we navigate through this paradigm shift in incorporating foundation models in robotics applications, it is imperative to recognize the challenges and potential risks that must be addressed in future research. Data scarcity in robotics applications, high variability in robotics settings, uncertainty quantification, safety evaluation, and real-time performance remain significant concerns that demand future research. We have delved into some of these challenges and have discussed potential avenues for improvement.

\section*{Acknowledgments}
The first author was supported on an ASEE e-Fellows postdoctoral fellowship. J.T. and S.T. were partially supported by NSF Graduate Research Fellowships.  This project was also partially supported by DARPA project HR001120C0107 and by a gift from Meta.  We are grateful for this support. Anirudha Majumdar was supported by the NSF CAREER Award [$\#$2044149] and the Office of Naval
Research [N00014-23-1-2148]. %
\bibliographystyle{unsrt}
\bibliography{main}

\end{document}